\documentclass[10pt,twocolumn,letterpaper]{article}

\usepackage[pagenumbers]{iccv} 

\DeclareMathOperator*{\argmax}{arg\,max}

\definecolor{iccvblue}{rgb}{0.21,0.49,0.74}
\usepackage[pagebackref,breaklinks,colorlinks,allcolors=iccvblue]{hyperref}
\usepackage{adjustbox}
\usepackage{xspace}
\usepackage{colortbl}
\usepackage{multirow}
\usepackage{pifont}
\usepackage{amsthm}
\usepackage[accsupp]{axessibility}
\newtheorem{proposition}{Proposition}

\def\paperID{284} 
\def\confName{ICCV}
\def\confYear{2025}

\title{Bidirectional Likelihood Estimation with \\ Multi-Modal Large Language Models for Text-Video Retrieval}

\author{
Dohwan Ko\textsuperscript{\rm 1}$^{*}$\hspace{0.4cm}
Ji Soo Lee\textsuperscript{\rm 1}$^{*}$\hspace{0.4cm}
Minhyuk Choi\textsuperscript{\rm 1}\hspace{0.4cm}
Zihang Meng\textsuperscript{\rm 2}\hspace{0.4cm}
Hyunwoo J. Kim\textsuperscript{\rm 3}$^{\dagger}$\vspace{0.3cm} \\
\textsuperscript{\rm 1}Korea University\hspace{0.8cm}
\textsuperscript{\rm 2}Meta GenAI\hspace{0.8cm}
\textsuperscript{\rm 3}KAIST\vspace{0.3cm} \\
\tt\small \{ikodoh, simplewhite9, sodlqnf123\}@korea.ac.kr\hspace{0.3cm}
\tt\small zihang@meta.com\hspace{0.3cm} hyunwoojkim@kaist.ac.kr
}

\begin{document}

\maketitle

\renewcommand{\thefootnote}{\fnsymbol{footnote}}
\footnotetext[0]{$*$ Equal contribution. $\dagger$ Corresponding authors.}

\begin{abstract}
    Text-Video Retrieval aims to find the most relevant text (or video) candidate given a video (or text) query from large-scale online databases.
    Recent work leverages multi-modal large language models (MLLMs) to improve retrieval, especially for long or complex query-candidate pairs.
    However, we observe that the naive application of MLLMs, i.e., retrieval based on candidate likelihood, introduces \textbf{candidate prior bias}, favoring candidates with inherently higher priors over those more relevant to the query.
    To this end, we propose a novel retrieval framework, Bidirectional Likelihood Estimation with MLLM (\textbf{BLiM}), which leverages both query and candidate likelihoods by training the model to generate text from a given video as well as video features from a given text.
    Furthermore, we introduce Candidate Prior Normalization (\textbf{CPN}), a simple yet effective training-free score calibration module designed to mitigate candidate prior bias in candidate likelihood.
    On four Text-Video Retrieval benchmarks, our BLiM equipped with CPN outperforms previous state-of-the-art models by 6.4 R@1 on average, effectively alleviating candidate prior bias and emphasizing query-candidate relevance.
    Our in-depth analysis across various multi-modal tasks beyond retrieval highlights the broad applicability of CPN which enhances visual understanding by reducing reliance on textual priors.
    Code is available at \href{https://github.com/mlvlab/BLiM}{https://github.com/mlvlab/BLiM}.
\end{abstract}
\section{Introduction}
\label{sec:intro}

\begin{figure*}[!t] 
    \centering
    \vspace{-7mm}
    \includegraphics[width=1.0\linewidth]{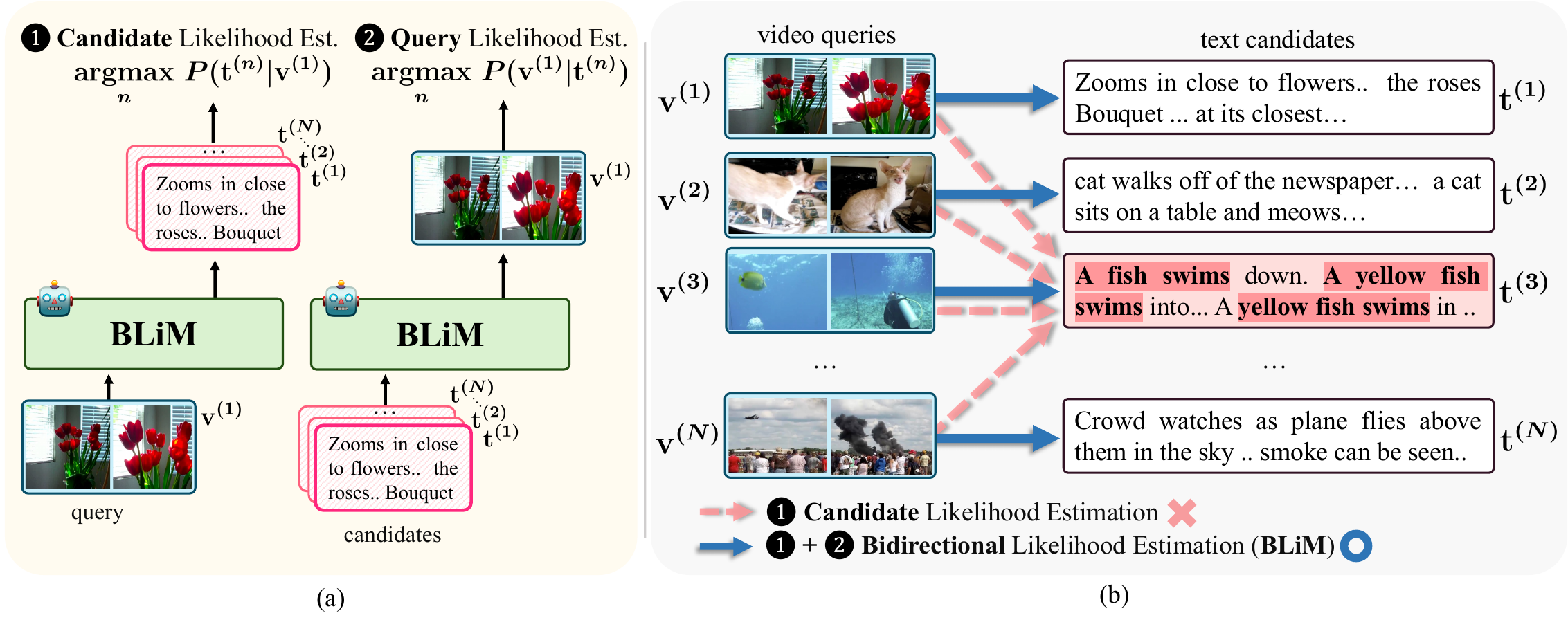}
    \caption{(a) provides an overview of BLiM for video-to-text retrieval, which leverages the bidirectional likelihood estimation with a query and candidate likelihoods to mitigate \textit{candidate prior bias}.
    (b) Candidate likelihood estimation tends to prioritize long and repetitive text with high prior probability.
    In contrast, bidirectional likelihood estimation of BLiM, effectively selects the most relevant text.}
    \label{fig:intro}
\end{figure*}
Text-Video Retrieval~\cite{lin2014microsoft,plummer2015flickr30k,xu2016msr,caba2015activitynet,anne2017localizing} aims to retrieve the most relevant text (or video) \textit{candidate} given a video (or text) \textit{query}.
To scale retrieval systems, previous works~\cite{luo2022clip4clip,xue2022clip} have primarily adopted dual-encoder architectures, leveraging encoder models such as BERT~\cite{devlin2018bert} and CLIP~\cite{radford2021learning}.
These models encode each query and candidate separately into single embeddings, enabling efficient retrieval via similarity between two embeddings.
While computationally efficient, its reliance on shallow similarity-based interactions restricts token-level alignment between queries and candidates, often leading to suboptimal retrieval performance.
To overcome this limitation, multi-modal large language models (MLLMs)-based~\cite{li2024videochat,Qwen2.5-VL,li2023mvbench,li2024llava,park2024llamo,wang2025internvideo2,lee2025vidchain,zhu2025internvl3,ko2025st} retrieval systems have been recently introduced~\cite{lin2024mm,chen2024mllm,liu2024lamra}.
Unlike dual-encoders, MLLM-based retrievers process concatenated query-candidate pairs, enabling deep token-level interactions, resulting in superior retrieval performance, particularly for long and complex query-candidate pairs.

However, we observe that naively maximizing candidate likelihood leads to \textit{candidate prior bias}, where candidates with higher prior probabilities are favored over those truly relevant to the query.
For instance, in Fig.~\ref{fig:intro}b, given a video query $\mathbf{v}$ and text candidates $\mathbf{t}$ in video-to-text retrieval, an MLLM retriever based on candidate likelihood $P(\mathbf{t}|\mathbf{v})$ tends to prioritize text candidates with frequently occurring patterns over those that are more semantically aligned with the video query.
In this example, such bias arises because MLLMs, due to their autoregressive nature, inherently assign higher probabilities to long and repetitive text, overlooking the actual content of the video query~\cite{wang2024mitigating}.
This prior bias is also prevalent in other multi-modal tasks, including visual question answering and captioning, where models tend to rely more on textual content than visual information when generating text responses~\cite{niu2021counterfactual,ramakrishnan2018overcoming,cadene2019rubi,leng2023mitigating}.
Similarly, in text-to-video retrieval, MLLMs often favor videos with static scenes over those exhibiting dynamic transitions.

To address candidate prior bias in MLLM-based retrieval systems, we propose a novel framework, Bidirectional Likelihood Estimation with MLLM (\textbf{BLiM}), which considers query likelihood as well as candidate likelihood.
Specifically, BLiM aims to generate text from a given video ($P(\mathbf{t}|\mathbf{v})$) and video features from a given text ($P(\mathbf{v}|\mathbf{t})$).
During inference, as in Fig.~\ref{fig:intro}, jointly considering both likelihoods allows BLiM to mitigate candidate prior bias by focusing on the semantic relevance between the query and candidate.
Additionally, we introduce Candidate Prior Normalization (\textbf{CPN}), a simple yet effective training-free score calibration module to reduce candidate prior bias in candidate likelihood estimation.
Equipped with CPN, BLiM achieves state-of-the-art performance by a remarkable margin on four popular Text-Video Retrieval benchmark datasets: DiDeMo~\cite{anne2017localizing}, ActivityNet~\cite{caba2015activitynet}, LSMDC~\cite{rohrbach2017movie}, and MSRVTT~\cite{xu2016msr}.
Furthermore, CPN enhances performance across various multi-modal tasks beyond retrieval by improving visual understanding through reduced reliance on textual priors, underscoring its broad applicability.

\noindent To sum up, our \textbf{contributions} are as follows:
\begin{itemize}
    \item[\textbullet] To the best of our knowledge, within the context of MLLMs for Text-Video Retrieval, this paper is the first to study the \textit{candidate prior bias} in candidate likelihood.
    \item[\textbullet] We propose BLiM, a novel MLLM-based retrieval system trained to generate text from video and video features from text, enabling bidirectional likelihood estimation.
    \item[\textbullet] We also present a simple yet effective score calibration module, CPN, which further reduces the candidate prior bias in candidate likelihood estimation.
    \item[\textbullet] Our BLiM, equipped with CPN, outperforms previous state-of-the-art models by an average margin of 6.4 in R@1, effectively alleviating candidate prior bias and emphasizing the relevance between the query and candidate.
\end{itemize}
\section{Candidate Prior Bias}
\label{sec:candidate}

We first analyze \textit{candidate prior bias} where retrieval using candidate likelihood of MLLMs heavily depends on the candidate prior probabilities, while ignoring actual query-candidate relevance.
The inference procedure of video-to-text retrieval using candidate likelihood is as follows:
\vspace{-1mm}
\begin{align}
    n^* & = \argmax_n \underbrace{P(\mathbf{t}^{(n)}|\mathbf{v})}_\text{candidate likelihood} = \argmax_n \frac{P(\mathbf{v}|\mathbf{t}^{(n)})P(\mathbf{t}^{(n)})}{P(\mathbf{v})} \nonumber \\
    & = \argmax_n \underbrace{P(\mathbf{v}|\mathbf{t}^{(n)})}_\text{query likelihood}\underbrace{P(\mathbf{t}^{(n)})}_\text{candidate prior}.
    \label{eq:v2t}
\end{align}
In Eq.~\eqref{eq:v2t}, the retrieval process is influenced by both the query likelihood $P(\mathbf{v}|\mathbf{t}^{(n)})$ and the candidate prior $P(\mathbf{t}^{(n)})$.
Ideally, retrieval should primarily rely on query likelihood to ensure semantic relevance between the query and the candidate.
However, since the candidate prior is independent of the query, it may introduce bias by prioritizing text candidates with frequently occurring patterns, even when they are less relevant to the given video query.
This bias leads to suboptimal retrieval, as in the following proposition:
\begin{proposition}
    Let $P(\mathbf{t}^{(m)}|\mathbf{v}^{(m)})$ denote the candidate likelihood for retrieving the most relevant text $\mathbf{t}^{(m)}$ given a query video $\mathbf{v}^{(m)}$. Suppose that:
    \begin{enumerate}
        \item The query likelihood correctly ranks $\mathbf{t}^{(m)}$ over any negative sample $\mathbf{t}^{(n)}$ and the gap is bounded as:
        \begin{align}
            0 < \log P(\mathbf{v}^{(m)}|\mathbf{t}^{(m)}) - \log P(\mathbf{v}^{(m)}|\mathbf{t}^{(n)}) < \varepsilon.
            \label{eq:query_gap}
        \end{align}
        \item There exists a text candidate $\mathbf{t}^{(n)}$ with a larger prior probability gap:
        \begin{align}
            \log P(\mathbf{t}^{(n)}) - \log P(\mathbf{t}^{(m)}) > c\varepsilon, \;\; \text{for some } c > 1.
            \label{eq:prior_gap}
        \end{align}
    \end{enumerate}
    Then, the candidate likelihood ranking is reversed:
    \begin{align}
        P(\mathbf{t}^{(m)}|\mathbf{v}^{(m)}) < P(\mathbf{t}^{(n)}|\mathbf{v}^{(m)}).
    \end{align}
\end{proposition}
\begin{proof}
    See Sec. D of the supplement.
\end{proof}

Fig.~\ref{fig:prior_bias} visualizes the impact of candidate prior bias. 
Interestingly, although query likelihood (Fig.~\ref{fig:prior_bias}b) yields relatively accurate retrieval results, the undesirable influence of candidate prior (Fig.~\ref{fig:prior_bias}c) distorts candidate likelihood estimation (Fig.~\ref{fig:prior_bias}a).
Specifically, the 24th text candidate (highlighted in a red box) exhibits the highest prior probability among all text candidates.
As a result, this text is retrieved for 374 out of 1,003 videos (37\%), when using candidate likelihood estimation, demonstrating that candidate likelihood is skewed by over-relying on the candidate prior.
Moreover, text-to-video retrieval follows a similar inference procedure, \ie, $n^* = \argmax_n P(\mathbf{v}^{(n)}|\mathbf{t}) = \argmax_n P(\mathbf{t}|\mathbf{v}^{(n)})P(\mathbf{v}^{(n)})$ and we find that candidate prior bias also exists in text-to-video retrieval, where the candidate likelihood $P(\mathbf{v}^{(n)}|\mathbf{t})$ overestimates the candidate prior $P(\mathbf{v}^{(n)})$, leading to the retrieval of irrelevant video candidates.
Further discussion on candidate prior bias in text-to-video retrieval is provided in Sec. E.1 and G.1 of the supplement.
These observations motivate us to consider both directions of likelihood, candidate and query likelihoods, to refine retrieval results by mitigating candidate prior bias in candidate likelihood estimation.

\begin{figure}[!t] 
    \centering
    \includegraphics[width=1.0\linewidth]{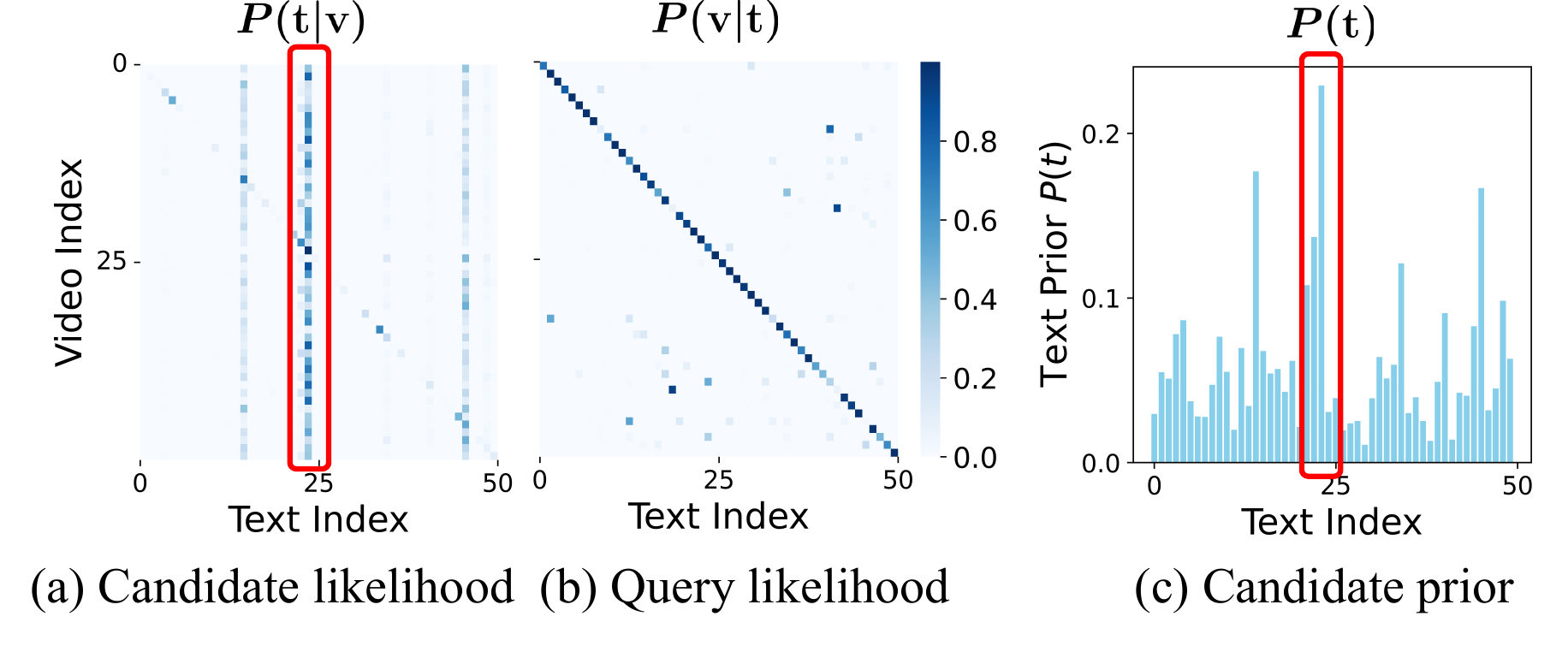}
    \caption{In video-to-text retrieval, similarities between queries and candidates using (a) candidate likelihood $P(\mathbf{t}|\mathbf{v})$ and (b) query likelihood $P(\mathbf{v}|\mathbf{t})$ are provided.
    (c) shows the candidate prior $P(\mathbf{t})$ for each text.
    To reduce visual clutter, 50 text-video pairs are sampled.
    Based on the candidate prior $P(\mathbf{t})$, the 24th text (highlighted in red) exhibits the highest prior probability in (c). 
    While the query videos correctly retrieve their corresponding text using query likelihood $P(\mathbf{v}|\mathbf{t})$ in (b), as indicated by the high similarity in diagonal elements, the text with the highest prior probability (red box) is frequently retrieved for irrelevant videos (374 out of 1,003) when using candidate likelihood $P(\mathbf{t}|\mathbf{v})$ in (a).
    }
    \label{fig:prior_bias}
\end{figure}
\section{Method}

Based on these observations, in Sec.~\ref{subsec:blim}, we propose Bidirectional Likelihood Estimation with MLLM (\textbf{BLiM}), a novel MLLM-based retrieval framework that incorporates both candidate and query likelihoods for Text-Video Retrieval.
Additionally, in Sec.~\ref{subsec:cpn}, we present a simple yet effective score calibration module, Candidate Prior Normalization (\textbf{CPN}), to mitigate candidate prior bias in the candidate likelihood estimation.

\begin{figure*}[!t] 
    \centering
    \includegraphics[width=1.0\linewidth]{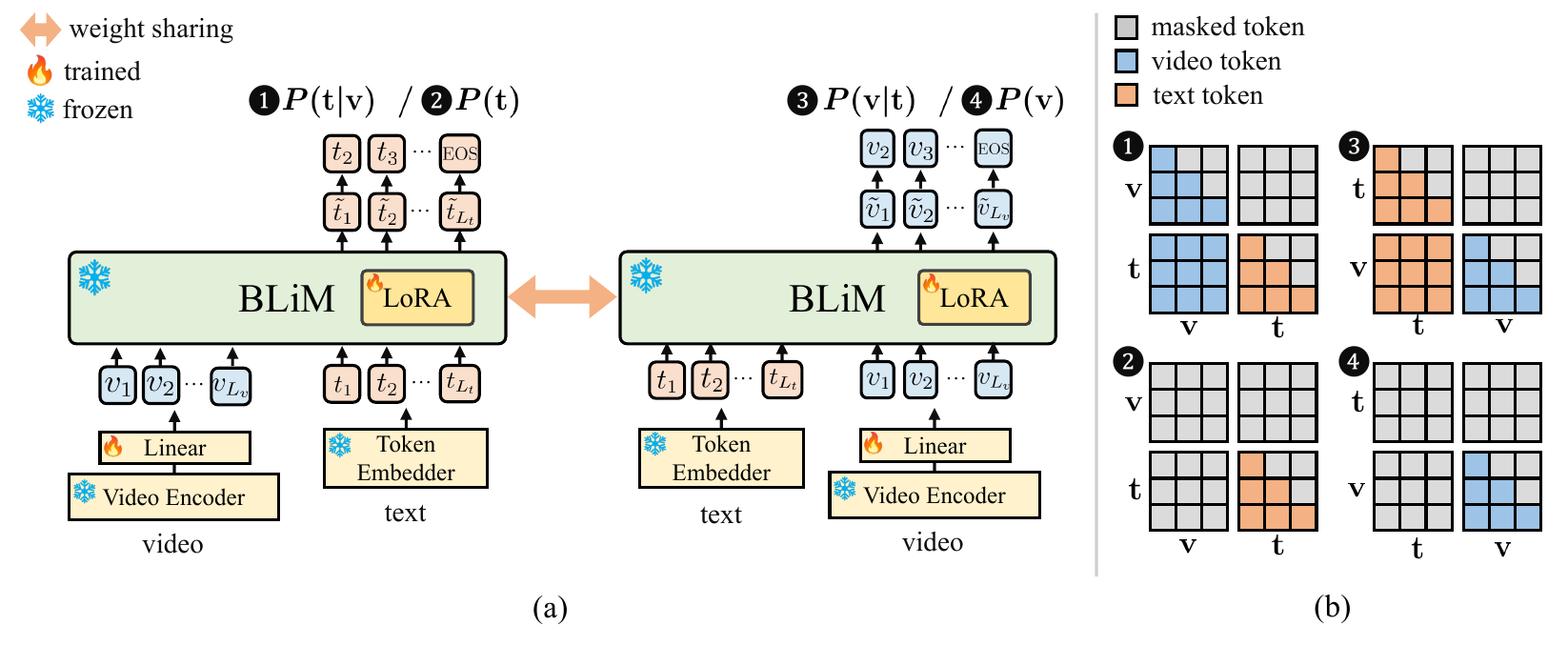}
    \caption{\textbf{Overall architecture.}
    (a) illustrates BLiM, which jointly maximizes both $P(\mathbf{t}|\mathbf{v})$ and $P(\mathbf{v}|\mathbf{t})$.
    (b) presents the attention masks used for estimating likelihoods and prior probabilities.
    To compute prior probabilities, attention masking is applied to all tokens of the input modality while generating the output modality.
    }
    \label{fig:main}
\end{figure*}
\subsection{Bidirectional Likelihood Estimation of MLLM}
\label{subsec:blim}

Unlike standard MLLMs usually trained to maximize $P(\mathbf{t}|\mathbf{v})$, we here propose BLiM, an MLLM that jointly maximizes bidirectional likelihoods $P(\mathbf{t}|\mathbf{v})$ and $P(\mathbf{v}|\mathbf{t})$.
The overall architecture of BLiM is depicted in Fig.~\ref{fig:main}a.

\noindent \textbf{Model architecture.}
BLiM is built upon the pretrained Video MLLM, VideoChat-Flash 7B~\cite{li2024videochat}, which consists of three key components: a video encoder (UMT~\cite{li2023unmasked}), a linear projection layer, and an LLM (Qwen2~\cite{yang2024qwen2}).
Given an input video, it is first segmented into $L_v$ clips, and the video encoder extracts visual features for each clip.
These features are then projected into the LLM's embedding space via the linear projection layer, forming visual tokens denoted as $\mathbf{v} = [v_1, \dots, v_{L_v}] \in \mathbb{R}^{L_v \times D}$, where $D$ is the hidden dimension.
These visual tokens are then concatenated with $L_t$ text tokens $\mathbf{t} = [t_1, \dots, t_{L_v}] \in \mathbb{R}^{L_t \times D}$ before being fed into the LLM.
We update only the linear projection layer and apply LoRA~\cite{hu2021lora} for parameter-efficient fine-tuning.

\noindent \textbf{Training procedure.}
BLiM is trained using a bidirectional likelihood maximization objective.
The first objective, video-grounded text generation $P(\mathbf{t}|\mathbf{v})$, follows the common pretraining paradigm of MLLMs as:
\begin{equation}
    \begin{split}
        \mathcal{L}_{t|v} = - \log P(\mathbf{t}|\mathbf{v}) & = - \sum_{i=1}^{L_t} \log P(t_i|t_{< i}, \mathbf{v}) \\
        & = \text{Softmax}(\text{Linear}(\tilde{t}_{i-1})),
    \end{split}
    \label{eq:vtg}
\end{equation}
where $\tilde{t}_{i-1} \in \mathbb{R}^D$ denotes the LLM's output representation corresponding to the $(i\!-\!1)$-th text token.

Additionally, we define a second objective, text-grounded video feature generation $P(\mathbf{v}|\mathbf{t})$, inspired by \cite{ko2023large}, as follows:
\begin{equation}
    \begin{split}
        \mathcal{L}_{v|t} & = - \log P(\mathbf{v}|\mathbf{t}) = -\sum_{i=1}^{L_v}\log P(v_i|v_{< i}, \mathbf{t}) \\
        & = - \sum_{i=1}^{L_v} \log \frac{\exp\left({\tilde{v}_{i-1}}^\top v_i\right)}{\sum_{n=1}^N\exp\left({\tilde{v}_{i-1}}^\top v_i^{(n)}\right)},
    \end{split}
    \label{eq:tvg}
\end{equation}
where $N$ is the number of videos in the training set and $\tilde{v}_{i-1} \in \mathbb{R}^D$ is the LLM's output representation of $v_{i-1} \in \mathbb{R}^D$.
Here, $v_0$ corresponds to the last token of the text sequence, allowing the model to generate video features conditioned on the entire text input.
In Eq.~\eqref{eq:tvg}, the model learns to autoregressively predict the next video clip feature $v_i$ given the preceding clips $v_{<i}$ and the text. 
The probability distribution is computed via a softmax function over candidate clips ${v_i^{(1)}, \dots, v_i^{(N)}}$, where the similarity score ${\tilde{v}_{i-1}}^\top v_i$ determines the likelihood of $v_i$ being the correct next clip.
This encourages the model to generate temporally coherent and text-consistent video representations.
Overall, we train BLiM by maximizing both likelihoods as:
\begin{equation}
    \mathcal{L}_\text{BLiM} = \mathcal{L}_{t|v} + \mathcal{L}_{v|t}.
    \label{eq:final}
\end{equation}
We note that the input modality order is swapped for each likelihood, as illustrated in Fig.~\ref{fig:main}a.
We use the prompt ``Describe this video.'' for $\mathcal{L}_{t|v}$ and ``Generate a video given the caption.'' for $\mathcal{L}_{v|t}$.

\noindent \textbf{Inference procedure.}
During inference, we combine both likelihoods $P(\mathbf{t}|\mathbf{v})$ and $P(\mathbf{v}|\mathbf{t})$ to find the most relevant candidate for a given query.
Given a video query and text candidates in video-to-text retrieval, $P(\mathbf{t}|\mathbf{v})$ and $P(\mathbf{v}|\mathbf{t})$ represent candidate likelihood and query likelihood, respectively, and the inference procedure is as follows:
\begin{equation}
    n_\text{V2T}^* = \argmax_n \underbrace{P(\mathbf{t}^{(n)}|\mathbf{v})}_\text{candidate likelihood} + \underbrace{P(\mathbf{v}|\mathbf{t}^{(n)})}_\text{query likelihood}.
    \label{eq:v2t_inference}
\end{equation}
On the other hand, in text-to-video retrieval with a text query and video candidates, the roles of likelihoods are reversed.
Then, the inference procedure is written as:
\begin{equation}
    n_\text{T2V}^* = \argmax_n \underbrace{P(\mathbf{t}|\mathbf{v}^{(n)})}_\text{query likelihood} + \underbrace{P(\mathbf{v}^{(n)}|\mathbf{t})}_\text{candidate likelihood}.
    \label{eq:t2v_inference}
\end{equation}

In both Eq.~\eqref{eq:v2t_inference} and \eqref{eq:t2v_inference}, candidate likelihood estimation identifies the candidate that is \textbf{most likely to be generated by the query}.
Conversely, query likelihood estimation identifies the candidate that is \textbf{most likely to generate the query}.
By jointly considering both likelihoods, BLiM mitigates the bias introduced by candidate prior probabilities, ensuring that the final prediction is based primarily on the semantic alignment between the query and the candidate.
Inference details are provided in Sec. C of the supplement.

Furthermore, to boost the retrieval efficiency, we adopt a two-stage retrieval pipeline~\cite{burges2006learning,nogueira2019passage,karpukhin2020dense,khattab2020colbert,li2023blip,li2023unmasked,wang2025internvideo2}, which first efficiently retrieves top-$K$ candidates with a small retrieval model and then refines the ranking with a large reranking model.
Specifically, we use InternVideo2 1B~\cite{wang2024internvideo2} to retrieve top-$K$ candidates and rerank their rankings using our BLiM.
This approach significantly reduces the inference time complexity from $O(N^2)$ to $O(KN)$ where $K \ll N$, resulting in more efficient inference than traversing all candidates (\eg, 307 times faster on ActivityNet).
\begin{table*}[!ht]
    \centering
    \begin{adjustbox}{width=\textwidth}
    \setlength{\tabcolsep}{2.0pt}
    \scriptsize
    \begin{tabular}{l|c c c|c c c|c c c|c c c|c c c|c c c|c c c|c c c}
        \toprule
        & \multicolumn{12}{c|}{\textbf{Text-to-Video}} & \multicolumn{12}{c}{\textbf{Video-to-Text}} \\
        & \multicolumn{3}{c|}{\textbf{DiDeMo}} & \multicolumn{3}{c|}{\textbf{ActivityNet}} & \multicolumn{3}{c|}{\textbf{LSMDC}} & \multicolumn{3}{c|}{\textbf{MSRVTT}} & \multicolumn{3}{c|}{\textbf{DiDeMo}}  & \multicolumn{3}{c|}{\textbf{ActivityNet}} & \multicolumn{3}{c|}{\textbf{LSMDC}} & \multicolumn{3}{c}{\textbf{MSRVTT}} \\
        & R@1 & R@5 & R@10 & R@1 & R@5 & R@10 & R@1 & R@5 & R@10 & R@1 & R@5 & R@10 & R@1 & R@5 & R@10 & R@1 & R@5 & R@10 & R@1 & R@5 & R@10 & R@1 & R@5 & R@10 \\
        \midrule
        \midrule
        \rowcolor[HTML]{FFF9C0}
        \multicolumn{25}{l}{\textbf{\textit{zero-shot}}} \\
        ViCLIP~\cite{wang2023internvid} & 18.4 & - & - & 15.1 & - & - & 20.1 & - & - & 42.4 & - & - & 27.9 & - & - & 24.0 & - & - & 16.9 & - & - & 41.3 & - & - \\
        InternVideo~\cite{wang2022internvideo} & 31.5 & - & - & 30.7 & - & - & 17.6 & - & - & 40.7 & - & - & 33.5 & - & - & 31.4 & - & - & 13.2 & - & - & 39.6 & - & - \\
        VideoCoCa~\cite{yan2022videococa} & - & - & - & 34.5 & 63.2 & 76.6 & - & - & - & 34.3 & 57.8 & 67.0 & - & - & - & 33.0 & 61.6 & 75.3 & - & - & - & 33.0 & 61.6 & 75.3 \\
        VideoPrism~\cite{zhao2024videoprism} & - & - & - & 52.7 & 79.4 & - & - & - & - & 52.7 & 77.2 & - & - & - & - & 50.3 & 77.1 & - & - & - & - & 51.7 & 75.2 & - \\
        UMT~\cite{li2023unmasked} & 48.6 & 72.9 & 79.0 & 41.9 & 68.9 & 80.3 & 24.9 & 41.7 & 51.8 & 40.7 & 63.4 & 71.8 & 49.9 & 74.8 & 81.4 & 39.4 & 66.8 & 78.3 & 21.9 & 37.8 & 45.7 & 37.1 & 58.7 & 68.9 \\
        InternVideo2 1B~\cite{wang2024internvideo2} & 57.0 & 80.0 & 85.1 & 60.4 & 83.9 & 90.8 & 32.0 & 52.4 & 59.4 & 51.9 & 75.3 & 82.5 & 54.3 & 77.2 & 83.5 & 54.8 & 81.5 & 89.5 & 27.3 & 44.2 & 51.6 & 50.9 & 73.4 & 81.8 \\
        InternVideo2 6B~\cite{wang2024internvideo2} & 57.9 & 80.0 & 84.6 & 63.2 & 85.6 & \textbf{92.5} & 33.8 & 55.9 & \textbf{62.2} & 55.9 & \textbf{78.3} & \textbf{85.1} & 57.1 & 79.9 & 85.0 & 56.5 & 82.8 & \textbf{90.3} & 30.1 & 47.7 & 54.8 & 53.7 & \textbf{77.5} & \textbf{84.1} \\
        \midrule
        \rowcolor[HTML]{BFF2FF}
        BLiM$^-$ (\textbf{Ours}) & \textbf{69.8} & \textbf{84.5} & \textbf{87.1} & \textbf{71.4} & \textbf{88.3} & 92.0 & \textbf{40.7} & \textbf{57.3} & 61.9 & \textbf{57.2} & 76.7 & 83.4 & \textbf{62.9} & \textbf{83.2} & \textbf{86.3} & \textbf{58.6} & \textbf{83.9} & 89.5 & \textbf{32.9} & \textbf{50.2} & \textbf{55.4} & \textbf{54.1} & 76.6 & \textbf{84.1} \\
        \midrule
        \rowcolor[HTML]{FFF9C0}
        \multicolumn{25}{l}{\textbf{\textit{fine-tuned}}} \\
        CLIP4Clip~\cite{luo2022clip4clip} & 42.8 & 68.5 & 79.2 &  40.5 & 72.4 & 83.4 &  21.6 & 41.8 & 49.8 & 44.5 & 71.4 & 81.6 & 42.5 & 70.6 & 80.2 &  42.6 & 73.4 & 85.6 &  20.9 & 40.7 & 49.1 & 43.1 & 70.5 & 81.2 \\
        ViCLIP~\cite{wang2023internvid} & 49.4 & - & - & 49.8 & - & - & 33.0 & - & - & 52.5 & - & - & 50.2 & - & - & 48.1 & - & - & 32.5 & - & - & 51.8 & - & - \\  
        MV-Adapter~\cite{jin2024mv} & 44.3 & 72.1 & 80.5 & 42.9 & 74.5 & 85.7 & 23.2 & 43.9 & 53.2 & 46.2 & 73.2 & 82.7 & 42.7 & 73.0 & 81.9 & 43.6 & 75.0 & 86.5 & 24.0 & 42.8 & 52.1 & 47.2 & 74.8 & 83.9 \\
        InternVideo~\cite{wang2022internvideo} & 57.9 & 82.4 & 88.9 & 62.2 & 85.9 & 93.2 & 34.0 & 53.7 & 62.9 & 55.2 & 79.6 & 87.5 & 59.1 & 81.8 & 89.0 & 62.8 & 86.2 & 93.3 & 34.9 & 54.6 & 63.1 & 57.9 & 79.2 & 86.4 \\
        UMT~\cite{li2023unmasked} & 70.4 & 90.1 & 93.5 & 66.8 & 89.1 & 94.9 & 43.0 & 65.5 & 73.0 & 58.8 & 81.0 & 87.1 & 67.9 & 88.6 & 93.0 & 64.4 & 89.1 & 94.8 & 41.4 & 64.3 & 71.5 & 58.6 & 81.6 & 86.5 \\
        Cap4Video~\citep{wu2023cap4video} & 52.0 & 79.4 & 87.5 &  - & - & - & - & - & - & 51.4 & 75.7 & 83.9 & - & - & - & - & - & - & - & - & - & 49.0 & 75.2 & 85.0 \\
        InternVideo2 1B$^*$~\cite{wang2024internvideo2} & 75.3 & 92.5 & 95.8 & 68.8 & 89.7 & 94.7 & 44.9 & 68.6 & 75.5 & 59.4 & 80.9 & 86.6 & 73.1 & 92.1 & 94.9 & 65.3 & 88.0 & 94.2 & 45.2 & 66.6 & 73.1 & 56.9 & 76.9 & 84.6 \\
        InternVideo2 6B~\cite{wang2024internvideo2} & 74.2 & - & - & 74.1 & - & - & 46.4 & - & - & 62.8 & - & - & 71.9 & - & - & 68.7 & - & - & 46.7 & - & - & 60.2 & - & - \\
        \midrule
        \rowcolor[HTML]{BFF2FF}
        BLiM (\textbf{Ours}) & \textbf{86.4} & \textbf{95.6} & \textbf{96.4} & \textbf{81.0} & \textbf{94.2} & \textbf{96.6} & \textbf{55.7} & \textbf{73.1} & \textbf{78.2} & \textbf{64.7} & \textbf{83.9} & \textbf{88.2} & \textbf{82.8} & \textbf{95.6} & \textbf{96.4} & \textbf{74.4} & \textbf{92.6} & \textbf{96.2} & \textbf{49.1} & \textbf{71.0} & \textbf{77.1} & \textbf{62.2} & \textbf{82.7} & \textbf{87.0} \\
        \bottomrule
    \end{tabular}
    \end{adjustbox}
    \caption{\textbf{Results on retrieval datasets.}
    $^-$ means that the prediction is performed without  $P(\mathbf{v}|\mathbf{t})$, and $^*$ denotes our reproduced results.
    }
    \label{tab:main}
\end{table*}
\subsection{Candidate Prior Normalization}
\label{subsec:cpn}

To further alleviate candidate prior bias in candidate likelihood, we here introduce a training-free score calibration module, CPN.
In video-to-text retrieval with text candidates, we aim to calibrate the candidate likelihood $P(\mathbf{t}|\mathbf{v})$ by normalizing the effect of the candidate prior $P(\mathbf{t})$ as:
\begin{equation}
    P(\mathbf{t}|\mathbf{v}) = \frac{P(\mathbf{v}|\mathbf{t})P(\mathbf{t})}{P(\mathbf{v})} \:\:\rightarrow\:\:\:
    \frac{P(\mathbf{t}|\mathbf{v})}{P(\mathbf{t})^\alpha} = \frac{P(\mathbf{v}|\mathbf{t})P(\mathbf{t})^{1-\alpha}}{P(\mathbf{v})},
    \label{eq:alpha}
\end{equation}
where $\alpha \in [0, 1]$ is a hyperparameter which determines a normalization strength.
Instead of directly using the standard candidate likelihood of $P(\mathbf{t}|\mathbf{v})$ in Eq.~(\ref{eq:alpha}) (left), we normalize it with the candidate prior $P(\mathbf{t})$ in Eq.~(\ref{eq:alpha}) (right).
When $\alpha=0$, the likelihood remains unchanged, while larger values of $\alpha$ apply stronger normalization to reduce the effect of the candidate prior.
Then, the normalized candidate likelihood $P^\alpha(\mathbf{t}|\mathbf{v})$ is defined as:
\begin{equation}
    \log P^{\alpha}(\mathbf{t}|\mathbf{v}) \triangleq \log \frac{P(\mathbf{t}|\mathbf{v})}{P(\mathbf{t})^{\alpha}} = \log P(\mathbf{t}|\mathbf{v}) - \alpha\log P(\mathbf{t}).
    \label{eq:prinorm}
\end{equation}
Also, in text-to-video retrieval with video candidates, the normalized candidate likelihood $P^\alpha(\mathbf{v}|\mathbf{t})$ is similarly defined to reduce the effect of the video candidate prior $P(\mathbf{v})$.
To calculate prior probabilities $P(\mathbf{t}) = \prod_iP(t_i|t_{< i})$ and $P(\mathbf{v}) = \prod_iP(v_i|v_{< i})$, attention masking, as illustrated in Fig.~\ref{fig:main}b, is applied to all tokens within the condition modality when predicting the other modality.
During inference, we use the normalized likelihood $P^\alpha(\mathbf{t}|\mathbf{v})$ and $P^\alpha(\mathbf{v}|\mathbf{t})$ to search for the optimal candidate in video-to-text and text-to-video retrievals, respectively. 
This reduces bias toward the candidate prior and leads to more balanced predictions. 
Specifically, we replace candidate likelihood $P(\mathbf{t}^{(n)}|\mathbf{v})$ in Eq.~\eqref{eq:v2t_inference} with $P^\alpha(\mathbf{t}^{(n)}|\mathbf{v})$ (similarly in Eq.~\eqref{eq:t2v_inference}).
The sensitivity study of $\alpha$ is available in Sec. F.2 of the supplement.

We also observe that the prior bias is prevalent in diverse multi-modal tasks.
Therefore, we extend CPN into a decoding scheme for a wide range of multi-modal tasks, \eg, visual question answering and visual captioning.
In these tasks, standard decoding introduces prior bias toward text, leading to hallucination problems due to ungrounded generation that neglects the visual content.
To mitigate this issue, instead of the standard decoding based on the likelihood $P(\mathbf{t}|\mathbf{v})$, we use normalized likelihood $P^\alpha(\mathbf{t}|\mathbf{v})$ to decode the text sequence.
By applying our normalized likelihood to various sampling strategies (\eg, nucleus sampling), the model generates a debiased text sequence, reducing the reliance on textual content and focusing more on visual content, thus minimizing hallucinations.
\section{Experiments}

In this section, we first showcase the result of BLiM on four popular Text-Video Retrieval benchmark datasets in Sec.~\ref{subsec:blim_result}.
We then verify the effectiveness of Bidirectional Likelihood Estimation in Sec.~\ref{subsec:ble}, and provide an extensive analysis of Candidate Prior Normalization in Sec.~\ref{subsec:cpn_analysis}.

\noindent \textbf{Datasets.}
For Text-Video Retrieval, we use DiDeMo~\citep{anne2017localizing}, ActivityNet~\citep{caba2015activitynet}, LSMDC~\citep{rohrbach2017movie}, and MSRVTT~\citep{xu2016msr} which contain diverse-length video and caption pairs.
We use the Recall@K (R@1, R@5, R@10) metric to evaluate the model's performance.

\noindent \textbf{Implementation details.}
An input video is divided into four clips, and each clip consists of four frames, resulting in a total of 16 frames per video.
During inference, we retrieve the top-16 candidates per query using InternVideo2 1B~\cite{wang2024internvideo2} and conduct a reranking among these candidates using our BLiM for accurate retrieval.
Further dataset and implementation details are in Sec. A and B of the supplement.

\subsection{Results of BLiM}
\label{subsec:blim_result}

\noindent \textbf{Comparison with state-of-the-art models.} 
In Tab.~\ref{tab:main}, we compare our model with state-of-the-art models on both text-to-video and video-to-text retrievals.
First, in the zero-shot setting, since pretrained MLLMs are typically trained to maximize $P(\mathbf{t}|\mathbf{v})$ and lack the ability to estimate $P(\mathbf{v}|\mathbf{t})$, retrieval is performed solely with $P(\mathbf{t}|\mathbf{v})$ with our CPN, denoted as BLiM$^-$.
Even without query likelihood estimation, BLiM$^-$ significantly outperforms previous state-of-the-art models, surpassing InternVideo2 6B by an average of 4.9 in R@1 across all datasets.
Moreover, with the bidirectional likelihood estimation in the fine-tuning setting, our BLiM achieves a new state-of-the-art performance on all benchmarks.
For example, on DiDeMo in text-to-video retrieval, BLiM improves R@1 by 12.2 compared to InternVideo2 6B.
As a result, the average R@1 gap between BLiM and InternVideo2 6B is 6.4.
Overall, BLiM achieves a remarkable performance gain both in zero-shot and fine-tuned settings, underscoring its effectiveness in Text-Video Retrieval.

\begin{table}[!t]
    \centering
    \begin{adjustbox}{width=\linewidth}
    \begin{tabular}{l|c|c c c c|c}
        \toprule
        & & BEiT-3~\cite{wang2022image} & ALBEF~\cite{li2021align} & BLIP~\cite{li2022blip} & BLIP-2~\cite{li2023blip} & \cellcolor[HTML]{BFF2FF}BLiM \\
        \midrule
        \midrule
        \multirow{2}{*}{COCO} & T2I & 65.1 & 60.7 & 65.1 & 68.3 & \cellcolor[HTML]{BFF2FF}\textbf{69.7}\\
        & I2T & 82.7 & 77.6 & 82.4 & \textbf{85.4} & \cellcolor[HTML]{BFF2FF}84.2 \\
        \midrule
        \multirow{2}{*}{Flickr30K} & T2I & 89.1 & 82.8 & 86.7 & 89.7 & \cellcolor[HTML]{BFF2FF}\textbf{92.1} \\
        & I2T & 97.5 & 94.1 & 96.7 & 97.6 & \cellcolor[HTML]{BFF2FF}\textbf{97.9} \\
        \bottomrule
    \end{tabular}
    \end{adjustbox}
    \caption{\textbf{Results in Text-Image Retrieval on Flickr30K and COCO.}
    We only report R@1 both in text-to-image (T2I) and image-to-text (I2T) retrieval.}
    \label{tab:image}
\end{table}

\begin{table}[!t]
    \centering
    \begin{adjustbox}{width=\linewidth}
    \begin{tabular}{l|c c|c c|c c|c c}
        \toprule
        & \multicolumn{2}{c|}{\textbf{DiDeMo}} & \multicolumn{2}{c|}{\textbf{ActivityNet}} & \multicolumn{2}{c|}{\textbf{LSMDC}} & \multicolumn{2}{c}{\textbf{MSRVTT}} \\
        & T2V & V2T & T2V & V2T & T2V & V2T & T2V & V2T \\
        \midrule
        \midrule
        MM-Embed~\cite{lin2024mm} & 81.6 & 79.7 & 78.5 & 70.7 & 52.8 & 48.1 & 61.2 & 60.5 \\
        RagVL~\cite{chen2024mllm} & 83.2 & 81.0 & 80.1 & 70.9 & 53.1 & 48.5 & 63.0 & 60.8 \\
        LamRA~\cite{liu2024lamra} & 83.4 & 79.2 & 76.0 & 68.7 & 51.9 & 47.8 & 59.7 & 60.7 \\
        \midrule
        \rowcolor[HTML]{BFF2FF}
        BLiM (\textbf{Ours}) & \textbf{86.4} & \textbf{82.8} & \textbf{81.0} & \textbf{74.4} & \textbf{55.7} & \textbf{49.1} & \textbf{64.7} & \textbf{62.2} \\
        \bottomrule
    \end{tabular}
    \end{adjustbox}
    \caption{\textbf{Comparison with other MLLM-based reranking methods.}
    We only report R@1 both in text-to-video (T2V) retrieval and video-to-text retrieval (V2T).
    }
    \label{tab:mllm}
\end{table}

\noindent \textbf{Extension to Text-Image Retrieval.}
We observe that the bidirectional likelihood estimation of BLiM can be generally applicable to other multi-modal retrieval tasks.
To validate its adaptability, we conduct experiments on Text-Image Retrieval by slightly modifying BLiM for image-based retrieval.
Specifically, instead of predicting a sequence of video clips for $P(\mathbf{v}|\mathbf{t})$, BLiM directly predicts a single image feature, while the estimation of $P(\mathbf{t}|\mathbf{v})$ remains unchanged.
Tab.~\ref{tab:image} presents results on Flickr30k~\cite{plummer2015flickr30k} and COCO~\cite{lin2014microsoft}. 
Notably, BLiM outperforms strong Text-Image Retrieval baselines, including BLIP-2~\cite{li2023blip}, achieving a new state-of-the-art performance in 3 out of 4 settings. 
For instance, in text-to-image retrieval on Flickr30k, R@1 is increased by 2.4 over BLIP-2, demonstrating the effectiveness of bidirectional likelihood estimation even in image-based retrieval tasks.

\noindent \textbf{Comparison with MLLM-based retrieval methods.}
We here compare BLiM with other MLLM-based retrieval methods~\cite{lin2024mm,chen2024mllm,liu2024lamra}.
Since MLLM-based retrievers have not been explored in the context of Text-Video Retrieval, we reproduce these algorithms in this setting.
For example, MM-Embed~\cite{lin2024mm}, prompts the MLLM to assess whether a query-candidate pair is semantically aligned by answering either ``True'' or ``False'' to the question: ``Does the video match the caption?''
The model then reranks candidates based on the logit of ``True.''
On the other hand, BLiM, equipped with CPN for candidate prior bias alleviation, directly estimates the likelihood $P(\mathbf{t}|\mathbf{v})$, capturing how likely the text is generated by the given video and vice versa.
For a fair comparison, we employ the same backbone MLLM (VideoChat-Flash) and apply reranking to the top-16 candidates per query retrieved by InternVideo2 1B across all methods.
As shown in Tab.~\ref{tab:mllm}, BLiM consistently outperforms other MLLM-based retrieval methods across all datasets, underscoring the advantages of using bidirectional likelihood estimation on MLLM-based retrieval.

\noindent \textbf{Discussion on computational cost.}
In Tab.~\ref{tab:cost}, we analyze the computational cost of BLiM by comparing its GPU memory usage and latency with a strong retrieval baseline, InternVideo2~\cite{wang2024internvideo2}.
BLiM, a 7B-parameter model, employs a two-stage retrieval process: it first retrieves the top-$K$ candidates using InternVideo2 1B, and then reranks them via bidirectional likelihood estimation.
As a result, its overall latency includes the retrieval time of InternVideo2 1B.
In text-to-video retrieval, BLiM improves the average R@1 by 7.6 over InternVideo2 6B, with only an additional 0.46 seconds required to process a single query, while consuming comparable GPU memory.
\begin{table}[t]
    \centering
    \setlength{\tabcolsep}{10.0pt}
    \begin{adjustbox}{width=\linewidth}
    \begin{tabular}{c|c|c|c c}
        \toprule
        \multirow{2}{*}{\textbf{Models}} & \multirow{2}{*}{\textbf{R@1}} & \textbf{GPU memory} & \multicolumn{2}{c}{\textbf{Latency} (seconds)} \\
        & & GB & Per query & Total \\
        \midrule
        \midrule
        InternVideo2-1B & 62.1 & \textbf{24} & \textbf{0.37} & \textbf{730.12} \\
        InternVideo2-6B & 64.4 & 27 & 1.29 & 2625.26 \\
        \midrule
        \rowcolor[HTML]{BFF2FF}
        BLiM-7B (\textbf{Ours}) & \textbf{72.0} & 27 & 1.75 & 3767.01 \\
        \bottomrule
    \end{tabular}
    \end{adjustbox}
    \caption{\textbf{Computational cost on text-to-video retrieval.}
    We report average results across four datasets. 
    Latency is measured using 8 $\times$ A6000 GPUs.
    }
    \label{tab:cost}
\end{table}
\begin{table}[!t]
    \centering
    \begin{adjustbox}{width=\linewidth}
    \begin{tabular}{l|c c c c|c}
        \toprule
        & \textbf{DiDeMo} & \textbf{ActivityNet} & \textbf{LSMDC} & \textbf{MSRVTT} & \textbf{avg.} \\
        \midrule
        \midrule
        CLE & 34.4 & 29.0 & 19.2 & 26.4 & 27.3 \\
        QLE & 72.5 & 69.5 & 43.7 & 56.4 & 60.5 \\
        \rowcolor[HTML]{BFF2FF}
        BLE (CLE + QLE) & \textbf{74.1} & \textbf{69.9} & \textbf{44.4} & \textbf{56.7} & \textbf{61.3} \\
        \bottomrule
    \end{tabular}
    \end{adjustbox}
    \caption{\textbf{Ablation study on bidirectional likelihood estimation.}
    We compare the performance of each likelihood estimation: candidate likelihood estimation (CLE), query likelihood estimation (QLE), and bidirectional likelihood estimation (BLE).
    We report the average R@1 for text-to-video and video-to-text retrieval, and exclude CPN in this experiment.
    }
    \label{tab:ble}
\end{table}

\begin{table*}[!t]
    \centering
    \begin{adjustbox}{width=0.99\textwidth}
    \begin{tabular}{l|c c c c|c c c c c c|c}
        \toprule
        & \multicolumn{4}{c|}{\textbf{Image Understanding Benchmark}} & \multicolumn{6}{c|}{\textbf{Video Understanding Benchmark}} & \\
        \cmidrule{2-11}
        \multicolumn{1}{c|}{\textbf{Model}} & \multicolumn{2}{c}{\textbf{MME}} & \textbf{MMBench} & \textbf{SeedBench} & \textbf{MVBench} & \multicolumn{2}{c}{\textbf{VideoMME}} & \textbf{MLVU} & \textbf{NExT-QA} & \textbf{SeedBench} & \multicolumn{1}{c}{\textbf{avg. $\Delta$}}\\
        \cmidrule{2-11}
        & perception & cognition & en-dev & image & test & w/o subtitle & w/ subtitle & m-avg & mc-val & video \\
        \midrule
        \midrule
        GPT-4V~\cite{achiam2023gpt} & 1409.0 & 517.0 & 75.0 & 49.9 & 43.5 & 59.9 & 63.3 & 49.2 & - & 60.5 & - \\
        VILA~\citep{lin2024vila} & \multicolumn{2}{c}{1762.0} & 82.4 & 75.8 & - & 60.1 & 61.1 & - & 67.9 & - & - \\
        IXC-2.5~\citep{zhang2024internlm} & \multicolumn{2}{c}{2229.0} & 82.2 & 75.4 & 69.1 & 55.8 & 58.8 & 37.3 & 71.0 & - & -\\
        \midrule
        VideoChat2~\citep{li2023mvbench} & 1231.4 & 274.3 & 63.9 & 67.8 & 60.1 & 42.2 & 53.0 & 45.8 & 78.9 & 54.5 & -\\
        \rowcolor[HTML]{BFF2FF}
        VideoChat2$^\dagger$ (\textbf{Ours}) & \textbf{1284.5} & \textbf{322.5} & \textbf{66.2} & \textbf{68.0} & \textbf{62.3} & \textbf{47.1} & \textbf{56.3} & \textbf{48.5} & \textbf{79.4} & \textbf{55.4} & \textbf{+11.8} \\
        \midrule
        LLaVA-Onevision~\citep{li2024llava} & 1696.7 & 514.6 & 79.8 & 75.0 & 57.1 & 58.5 & 57.8 & 65.3 & 79.4 & 56.9 & -\\
        \rowcolor[HTML]{BFF2FF}
        LLaVA-Onevision$^\dagger$ (\textbf{Ours}) & \textbf{1708.6} & \textbf{535.0} & \textbf{81.3} & \textbf{75.3} & \textbf{58.9} & \textbf{61.7} & \textbf{62.1} & \textbf{65.8} & \textbf{79.5} & \textbf{57.0} & \textbf{+4.4} \\
        \midrule
        InternVL2~\citep{chen2024far} & 1622.7 & 582.5 & 81.8 & 76.1 & 65.8 & 51.3 & 51.7 & 50.8 & 80.4 & 56.4 & - \\
        \rowcolor[HTML]{BFF2FF}
        InternVL2$^\dagger$ (\textbf{Ours}) & \textbf{1642.1} & \textbf{590.0} & \textbf{82.7} & \textbf{76.2} & \textbf{67.1} & \textbf{54.7} & \textbf{55.1} & \textbf{55.1} & \textbf{80.8} & \textbf{56.6} & \textbf{+4.1} \\
        \bottomrule
    \end{tabular}
    \end{adjustbox}
    \caption{\textbf{Results of CPN decoding.}
    The performances on seven different benchmarks are reported.
    $\dagger$ means the model with CPN decoding.
    }
    \label{tab:video_benchmark}
\end{table*}
\begin{figure}[!t] 
    \centering
    \includegraphics[width=1.0\linewidth]{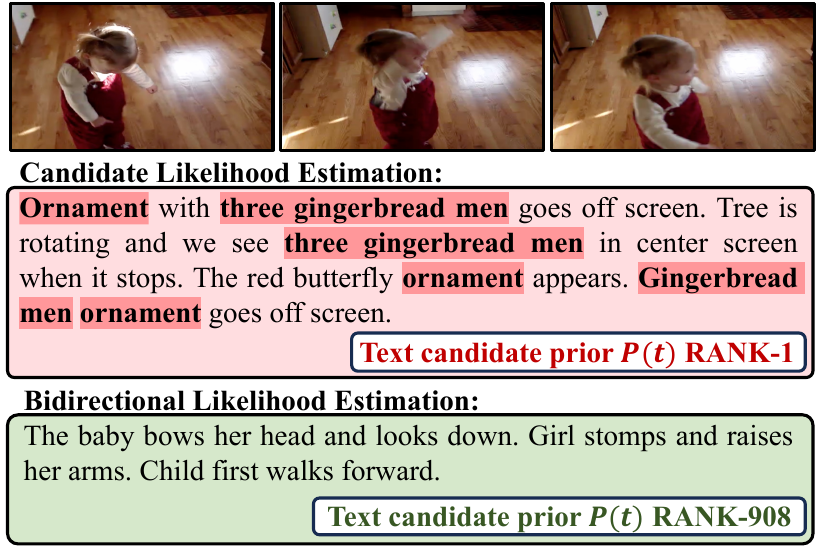}
    \caption{\textbf{A retrieval example in video-to-text retrieval on DiDeMo.}
    Green indicates a correct prediction, while red denotes an incorrect one.
    Repeated phrases are highlighted in red.
    }
    \label{fig:ble}
\end{figure}
\subsection{Analysis of Bidirectional Likelihood Estimation}
\label{subsec:ble}

\noindent \textbf{Quantitative analysis.}
In Tab.~\ref{tab:ble}, we conduct an ablation study on bidirectional likelihood estimation of BLiM to verify its effectiveness in alleviating candidate prior bias.
To isolate the impact of bidirectional likelihood estimation, we exclude CPN from this analysis.
Across all datasets, candidate likelihood estimation (CLE) is highly susceptible to candidate prior bias, leading to suboptimal retrieval performance, whereas query likelihood estimation (QLE) achieves notable improvements over candidate likelihood estimation by alleviating such bias.
Specifically, R@1 is improved by 38.1, 40.5, 24.5, and 30.0 on DiDeMo, ActivityNet, LSMDC, and MSRVTT, respectively.
Moreover, bidirectional likelihood estimation (BCE) further enhances performance over query likelihood estimation alone, \eg, 1.6 R@1 gain on DiDeMo.
As a result, the integration of query likelihood estimation is pivotal in mitigating candidate prior bias in model predictions.
Detailed results for both text-to-video and video-to-text retrieval tasks are presented in Sec. F.3 of the supplement.

\noindent \textbf{Qualitative analysis.}
We provide a qualitative example in Fig.~\ref{fig:ble} to show the impact of bidirectional likelihood estimation.
We observe that bidirectional likelihood estimation successfully retrieves the ground-truth text from the given video, while the candidate likelihood estimation tends to retrieve incorrect text that disregards the video content.
Notably, the ground-truth text ranks 908 out of 1,003 based on candidate prior probability, while the incorrect text predicted by candidate likelihood estimation holds the highest prior probability (ranked 1).
We also find that texts with high candidate prior probabilities tend to be longer and contain repetitive phrases (\eg, ``ornament'' and ``gingerbread men'') due to the autoregressive nature of LLMs.
Surprisingly, the correlation between prior probabilities and the text length is 0.97, and the correlation between prior probabilities and the number of repetitive phrases is 0.93 (see Sec. E.3).
Overall, our analysis underscores that high text candidate prior probability can hinder accurate retrieval, as it leads to a preference for common or verbose texts rather than contextually appropriate ones.
A similar trend is observed in text-to-video retrieval, where candidate likelihood estimation tends to prefer high-prior videos that often contain static scenes with limited temporal dynamics (see Sec. E.1).
In contrast, our bidirectional likelihood estimation approach mitigates this bias by prioritizing semantic alignment over statistical frequency.
\begin{table}[!t]
    \centering
    \begin{adjustbox}{width=\linewidth}
    \begin{tabular}{l|c|c c c c|c}
        \toprule
        & \textbf{CPN} & \textbf{DiDeMo} & \textbf{ActivityNet} & \textbf{LSMDC} & \textbf{MSRVTT} & \textbf{avg. $\Delta$} \\
        \midrule
        \midrule
        CLE & \ding{56} & 34.4 & 29.0 & 19.2 & 26.4 & - \\
        \rowcolor[HTML]{BFF2FF}
        CLE & \ding{52} & \textbf{59.2} & \textbf{46.3} & \textbf{31.7} & \textbf{44.3} & \textbf{+18.1} \\
        \midrule
        BLE & \ding{56} & 74.1 & 69.9 & 44.4 & 56.7 & - \\
        \rowcolor[HTML]{BFF2FF}
        BLE & \ding{52} & \textbf{81.3} & \textbf{73.7} & \textbf{47.6} & \textbf{59.3} & \textbf{+4.2} \\
        \bottomrule
    \end{tabular}
    \end{adjustbox}
    \caption{\textbf{Ablation study on CPN.}
    The average R@1 is reported.
    }
    \label{tab:cpn}
\end{table}
\subsection{Analysis of Candidate Prior Normalization}
\label{subsec:cpn_analysis}

\noindent \textbf{Abaltion study on CPN.}
Tab.~\ref{tab:cpn} demonstrates an ablation study on CPN in Text-Video Retrieval.
We observe a substantial performance improvement after applying CPN to candidate likelihood estimation, with R@1 gains of 24.8, 17.3, 12.5, and 17.9 on each dataset.
Consequently, incorporating CPN leads to an average R@1 improvement of 4.2 in bidirectional likelihood estimation. 
These findings suggest that CPN serves as a simple yet effective plug-and-play module for mitigating candidate prior bias.
Detailed results for both text-to-video and video-to-text retrieval tasks are presented in Sec. F.4 of the supplement.

\noindent \textbf{CPN decoding on various multi-modal benchmarks.}
We present an in-depth analysis of CPN decoding on multi-modal understanding benchmarks beyond mere retrieval tasks.
Tab.~\ref{tab:video_benchmark} presents evaluation results on seven image and video understanding benchmarks (MME~\citep{fu2023mme}, MMBench~\citep{liu2023mmbench}, SeedBench~\citep{li2023seed}, MVBench~\citep{li2023mvbench}, VideoMME~\citep{fu2024video}, MLVU~\citep{zhou2024mlvu}, and NExT-QA~\citep{xiao2021next}) encompassing comprehensive tasks that assess the model's image and video understanding as well as reasoning abilities.
By applying CPN decoding to three different MLLMs (VideoChat2~\citep{li2023mvbench}, LLaVA-Onevision~\citep{li2024llava}, and InternVL2~\citep{chen2024far}), the performances are consistently improved across all the benchmarks by average margins of 11.8, 4.4, and 4.1, respectively.
This result indicates that our training-free score calibration method not only enhances retrieval but also significantly boosts the overall reasoning and comprehension capabilities of the models.

\begin{figure}[!t]
    \centering
    \includegraphics[width=1.0\linewidth]{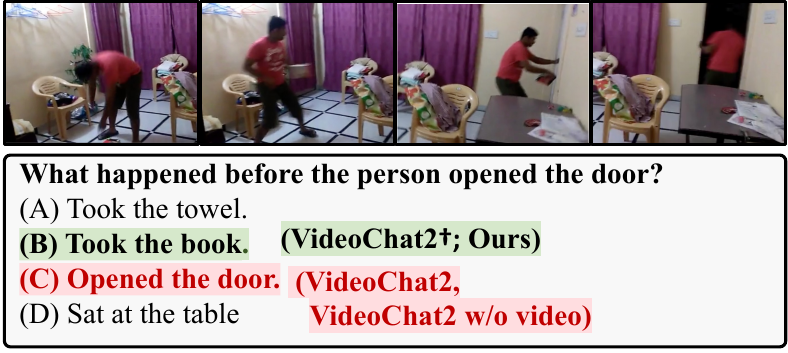}
    \caption{\textbf{A qualitative example of CPN decoding on MVBench.}
    Green signifies the accurate prediction, while red denotes the incorrect prediction.
    $\dagger$ indicates the model with CPN decoding.
    }
    \label{fig:cpn_decoding}
\end{figure}

To illustrate how CPN decoding corrects the model's output, we provide qualitative results in Fig.~\ref{fig:cpn_decoding}, which compares the predictions of VideoChat2, VideoChat2 w/o video, and VideoChat2 + CPN decoding (\ie, VideoChat2$^\dagger$).
We note that the VideoChat2 w/o video model relies solely on textual information, \ie, text priors, for prediction.
We find that standard VideoChat2 often adheres to predictions based on the text prior (VideoChat2 w/o video), resulting in incorrect answers.
For the question, ``What happened before the person opened the door?'', the VideoChat2 w/o video model assigns high text prior probability to the option ``(C) Opened the door'' due to the repetition of the phrase in the question.
Thus, the standard VideoChat2's over-reliance on the wrong text prior results in inaccurate outputs, while our CPN decoding successfully mitigates this bias by encouraging the model to refer more to visual information.
Overall, CPN decoding is both model- and task-agnostic, serving as an effective score calibration module that reduces reliance on linguistic cues and ensures a more balanced consideration of visual and textual information for accurate predictions.
We provide an analysis of CPN decoding in visual captioning in Section E.2 of the supplement, highlighting its effectiveness in enhancing generation quality by reducing hallucination problems.
\section{Related Works}
\noindent \textbf{Text-Video Retrieval.}
Text-Video Retrieval is a widely studied multi-modal task that aims to find the most relevant video based on a text query or vice versa.
Early studies~\cite{luo2022clip4clip,gorti2022x,liu2022ts2,wu2023cap4video,xue2022clip} have leveraged CLIP~\cite{radford2021learning}, a dual-encoder architecture trained with contrastive loss to learn a shared embedding space between images and text, extending its text-image representations to the text-video domain.
For instance, CLIP4Clip~\cite{luo2022clip4clip} introduced image aggregation modules on top of CLIP to enhance temporal understanding in the video domain.
Another line of research has explored video foundation models such as UMT~\cite{li2023unmasked}, InternVideo~\cite{wang2022internvideo}, and InternVideo2~\cite{wang2024internvideo2}, which are trained on large-scale text-video datasets, achieving strong retrieval performance.
More recently, Cap4Video~\cite{wu2023cap4video} proposed utilizing auxiliary data, \eg, video captions, to enrich video representations by bridging the modality gap.

\noindent \textbf{MLLM-based retrieval systems.}
With the emergence of multi-modal large language models (MLLMs) demonstrating impressive performance in diverse multi-modal understanding tasks, recent studies~\cite{lin2024mm,chen2024mllm,liu2024lamra,lin2023revisiting,lin2024evaluating} have explored their application in multi-modal retrieval.
Unlike traditional dual-encoder architectures that rely on shallow similarity-based interactions between text and video, MLLMs enable fine-grained token-level interactions, capturing deeper semantic relationships.
For example, MM-Embed~\cite{lin2024mm} prompts an MLLM to evaluate the semantic alignment between a given query and candidate by assessing the logits of ``True'' in response to the question, ``Does the image match the caption?''.
Closely related to our work, VisualGPTScore~\cite{lin2023revisiting} investigates the influence of language priors in retrieval and introduces debiasing strategies to reduce their effect.
In this work, we observe that MLLM-based retrievers tend to favor candidates with higher prior probabilities rather than those most relevant to the query. 
To address this issue, we propose bidirectional likelihood estimation and candidate prior normalization to mitigate bias and improve retrieval accuracy.
\section{Conclusion}
In this paper, we observe that candidate likelihood estimation using MLLMs in Text-Video Retrieval tends to retrieve incorrect text from a given video (and vice versa) due to candidate prior bias.
To address this over-reliance on candidate priors, we propose Bidirectional Likelihood Estimation with MLLM (BLiM), which considers both candidate and query likelihoods.
Additionally, our simple plug-and-play score calibration module, Candidate Prior Normalization (CPN), further enhances performance alongside BLiM in Text-Video Retrieval by reducing dependence on candidate priors.
Our experimental results demonstrate the effectiveness of CPN decoding applied to MLLMs, facilitating a more balanced consideration of both textual and visual information across various multi-modal tasks.

\noindent \textbf{Acknowledgments.}
This work was partly supported by IITP grant funded by MSIP \& MSIT (No. RS-2024-00443251, No. RS-2024-00457882), NRF grant funded by MSIT (NRF-2023R1A2C2005373), and IITP-ITRC grant funded by MSIT (IITP-2025-RS-2024-00436857).

{
    \small
    \bibliographystyle{unsrt}
    \bibliography{main}

\begin{thebibliography}{10}

\bibitem{lin2014microsoft}
Tsung-Yi Lin, Michael Maire, Serge Belongie, James Hays, Pietro Perona, Deva Ramanan, Piotr Doll{\'a}r, and C~Lawrence Zitnick.
\newblock Microsoft coco: Common objects in context.
\newblock In {\em ECCV}, 2014.

\bibitem{plummer2015flickr30k}
Bryan~A Plummer, Liwei Wang, Chris~M Cervantes, Juan~C Caicedo, Julia Hockenmaier, and Svetlana Lazebnik.
\newblock Flickr30k entities: Collecting region-to-phrase correspondences for richer image-to-sentence models.
\newblock In {\em ICCV}, 2015.

\bibitem{xu2016msr}
Jun Xu, Tao Mei, Ting Yao, and Yong Rui.
\newblock Msr-vtt: A large video description dataset for bridging video and language.
\newblock In {\em CVPR}, 2016.

\bibitem{caba2015activitynet}
Fabian Caba~Heilbron, Victor Escorcia, Bernard Ghanem, and Juan Carlos~Niebles.
\newblock Activitynet: A large-scale video benchmark for human activity understanding.
\newblock In {\em CVPR}, 2015.

\bibitem{anne2017localizing}
Lisa Anne~Hendricks, Oliver Wang, Eli Shechtman, Josef Sivic, Trevor Darrell, and Bryan Russell.
\newblock Localizing moments in video with natural language.
\newblock In {\em ICCV}, 2017.

\bibitem{luo2022clip4clip}
Huaishao Luo, Lei Ji, Ming Zhong, Yang Chen, Wen Lei, Nan Duan, and Tianrui Li.
\newblock Clip4clip: An empirical study of clip for end to end video clip retrieval and captioning.
\newblock {\em arXiv preprint arXiv:2104.08860}, 2022.

\bibitem{xue2022clip}
Hongwei Xue, Yuchong Sun, Bei Liu, Jianlong Fu, Ruihua Song, Houqiang Li, and Jiebo Luo.
\newblock Clip-vip: Adapting pre-trained image-text model to video-language representation alignment.
\newblock In {\em ICLR}, 2023.

\bibitem{devlin2018bert}
Jacob Devlin.
\newblock Bert: Pre-training of deep bidirectional transformers for language understanding.
\newblock In {\em NAACL}, 2019.

\bibitem{radford2021learning}
Alec Radford, Jong~Wook Kim, Chris Hallacy, Aditya Ramesh, Gabriel Goh, Sandhini Agarwal, Girish Sastry, Amanda Askell, Pamela Mishkin, Jack Clark, et~al.
\newblock Learning transferable visual models from natural language supervision.
\newblock In {\em ICML}, 2021.

\bibitem{li2024videochat}
Xinhao Li, Yi~Wang, Jiashuo Yu, Xiangyu Zeng, Yuhan Zhu, Haian Huang, Jianfei Gao, Kunchang Li, Yinan He, Chenting Wang, et~al.
\newblock Videochat-flash: Hierarchical compression for long-context video modeling.
\newblock {\em arXiv preprint arXiv:2501.00574}, 2024.

\bibitem{Qwen2.5-VL}
Qwen Team.
\newblock Qwen2.5-vl, January 2025.

\bibitem{li2023mvbench}
Kunchang Li, Yali Wang, Yinan He, Yizhuo Li, Yi~Wang, Yi~Liu, Zun Wang, Jilan Xu, Guo Chen, Ping Luo, et~al.
\newblock Mvbench: A comprehensive multi-modal video understanding benchmark.
\newblock {\em CVPR}, 2024.

\bibitem{li2024llava}
Bo~Li, Yuanhan Zhang, Dong Guo, Renrui Zhang, Feng Li, Hao Zhang, Kaichen Zhang, Yanwei Li, Ziwei Liu, and Chunyuan Li.
\newblock Llava-onevision: Easy visual task transfer.
\newblock {\em arXiv preprint arXiv:2408.03326}, 2024.

\bibitem{park2024llamo}
Jinyoung Park, Minseong Bae, Dohwan Ko, and Hyunwoo~J Kim.
\newblock Llamo: Large language model-based molecular graph assistant.
\newblock In {\em NeurIPS}, 2024.

\bibitem{wang2025internvideo2}
Yi~Wang, Xinhao Li, Ziang Yan, Yinan He, Jiashuo Yu, Xiangyu Zeng, Chenting Wang, Changlian Ma, Haian Huang, Jianfei Gao, et~al.
\newblock Internvideo2. 5: Empowering video mllms with long and rich context modeling.
\newblock {\em arXiv preprint arXiv:2501.12386}, 2025.

\bibitem{lee2025vidchain}
Ji~Soo Lee, Jongha Kim, Jeehye Na, Jinyoung Park, and Hyunwoo~J Kim.
\newblock Vidchain: Chain-of-tasks with metric-based direct preference optimization for dense video captioning.
\newblock In {\em AAAI}, 2025.

\bibitem{zhu2025internvl3}
Jinguo Zhu, Weiyun Wang, Zhe Chen, Zhaoyang Liu, Shenglong Ye, Lixin Gu, Hao Tian, Yuchen Duan, Weijie Su, Jie Shao, et~al.
\newblock Internvl3: Exploring advanced training and test-time recipes for open-source multimodal models.
\newblock {\em arXiv preprint arXiv:2504.10479}, 2025.

\bibitem{ko2025st}
Dohwan Ko, Sihyeon Kim, Yumin Suh, Minseo Yoon, Manmohan Chandraker, Hyunwoo~J Kim, et~al.
\newblock St-vlm: Kinematic instruction tuning for spatio-temporal reasoning in vision-language models.
\newblock {\em arXiv preprint arXiv:2503.19355}, 2025.

\bibitem{lin2024mm}
Sheng-Chieh Lin, Chankyu Lee, Mohammad Shoeybi, Jimmy Lin, Bryan Catanzaro, and Wei Ping.
\newblock Mm-embed: Universal multimodal retrieval with multimodal llms.
\newblock In {\em ICLR}, 2025.

\bibitem{chen2024mllm}
Zhanpeng Chen, Chengjin Xu, Yiyan Qi, and Jian Guo.
\newblock Mllm is a strong reranker: Advancing multimodal retrieval-augmented generation via knowledge-enhanced reranking and noise-injected training.
\newblock {\em arXiv preprint arXiv:2407.21439}, 2024.

\bibitem{liu2024lamra}
Yikun Liu, Pingan Chen, Jiayin Cai, Xiaolong Jiang, Yao Hu, Jiangchao Yao, Yanfeng Wang, and Weidi Xie.
\newblock Lamra: Large multimodal model as your advanced retrieval assistant.
\newblock {\em arXiv preprint arXiv:2412.01720}, 2024.

\bibitem{wang2024mitigating}
Weichuan Wang, Zhaoyi Li, Defu Lian, Chen Ma, Linqi Song, and Ying Wei.
\newblock Mitigating the language mismatch and repetition issues in llm-based machine translation via model editing.
\newblock In {\em EMNLP}, 2024.

\bibitem{niu2021counterfactual}
Yulei Niu, Kaihua Tang, Hanwang Zhang, Zhiwu Lu, Xian-Sheng Hua, and Ji-Rong Wen.
\newblock Counterfactual vqa: A cause-effect look at language bias.
\newblock In {\em CVPR}, 2021.

\bibitem{ramakrishnan2018overcoming}
Sainandan Ramakrishnan, Aishwarya Agrawal, and Stefan Lee.
\newblock Overcoming language priors in visual question answering with adversarial regularization.
\newblock In {\em NeurIPS}, 2018.

\bibitem{cadene2019rubi}
Remi Cadene, Corentin Dancette, Matthieu Cord, Devi Parikh, et~al.
\newblock Rubi: Reducing unimodal biases for visual question answering.
\newblock In {\em NeurIPS}, 2019.

\bibitem{leng2023mitigating}
Sicong Leng, Hang Zhang, Guanzheng Chen, Xin Li, Shijian Lu, Chunyan Miao, and Lidong Bing.
\newblock Mitigating object hallucinations in large vision-language models through visual contrastive decoding.
\newblock In {\em CVPR}, 2024.

\bibitem{rohrbach2017movie}
Anna Rohrbach, Atousa Torabi, Marcus Rohrbach, Niket Tandon, Christopher Pal, Hugo Larochelle, Aaron Courville, and Bernt Schiele.
\newblock Movie description.
\newblock {\em IJCV}, 2017.

\bibitem{li2023unmasked}
Kunchang Li, Yali Wang, Yizhuo Li, Yi~Wang, Yinan He, Limin Wang, and Yu~Qiao.
\newblock Unmasked teacher: Towards training-efficient video foundation models.
\newblock In {\em ICCV}, 2023.

\bibitem{yang2024qwen2}
An~Yang, Baosong Yang, Beichen Zhang, Binyuan Hui, Bo~Zheng, Bowen Yu, Chengyuan Li, Dayiheng Liu, Fei Huang, Haoran Wei, et~al.
\newblock Qwen2. 5 technical report.
\newblock {\em arXiv preprint arXiv:2412.15115}, 2024.

\bibitem{hu2021lora}
Edward~J Hu, Yelong Shen, Phillip Wallis, Zeyuan Allen-Zhu, Yuanzhi Li, Shean Wang, Lu~Wang, and Weizhu Chen.
\newblock Lora: Low-rank adaptation of large language models.
\newblock In {\em ICLR}, 2022.

\bibitem{ko2023large}
Dohwan Ko, Ji~Soo Lee, Wooyoung Kang, Byungseok Roh, and Hyunwoo~J Kim.
\newblock Large language models are temporal and causal reasoners for video question answering.
\newblock In {\em EMNLP}, 2023.

\bibitem{burges2006learning}
Christopher Burges, Robert Ragno, and Quoc Le.
\newblock Learning to rank with nonsmooth cost functions.
\newblock In {\em NeurIPS}, 2006.

\bibitem{nogueira2019passage}
Rodrigo Nogueira and Kyunghyun Cho.
\newblock Passage re-ranking with bert.
\newblock {\em arXiv preprint arXiv:1901.04085}, 2019.

\bibitem{karpukhin2020dense}
Vladimir Karpukhin, Barlas O{\u{g}}uz, Sewon Min, Patrick Lewis, Ledell Wu, Sergey Edunov, Danqi Chen, and Wen-tau Yih.
\newblock Dense passage retrieval for open-domain question answering.
\newblock In {\em EMNLP}, 2020.

\bibitem{khattab2020colbert}
Omar Khattab and Matei Zaharia.
\newblock Colbert: Efficient and effective passage search via contextualized late interaction over bert.
\newblock In {\em SIGIR}, 2020.

\bibitem{li2023blip}
Junnan Li, Dongxu Li, Silvio Savarese, and Steven Hoi.
\newblock Blip-2: Bootstrapping language-image pre-training with frozen image encoders and large language models.
\newblock In {\em ICML}, 2023.

\bibitem{wang2024internvideo2}
Yi~Wang, Kunchang Li, Xinhao Li, Jiashuo Yu, Yinan He, Guo Chen, Baoqi Pei, Rongkun Zheng, Zun Wang, Yansong Shi, et~al.
\newblock Internvideo2: Scaling foundation models for multimodal video understanding.
\newblock In {\em ECCV}, 2024.

\bibitem{wang2023internvid}
Yi~Wang, Yinan He, Yizhuo Li, Kunchang Li, Jiashuo Yu, Xin Ma, Xinhao Li, Guo Chen, Xinyuan Chen, Yaohui Wang, et~al.
\newblock Internvid: A large-scale video-text dataset for multimodal understanding and generation.
\newblock In {\em ICLR}, 2024.

\bibitem{wang2022internvideo}
Yi~Wang, Kunchang Li, Yizhuo Li, Yinan He, Bingkun Huang, Zhiyu Zhao, Hongjie Zhang, Jilan Xu, Yi~Liu, Zun Wang, et~al.
\newblock Internvideo: General video foundation models via generative and discriminative learning.
\newblock {\em arXiv preprint arXiv:2212.03191}, 2022.

\bibitem{yan2022videococa}
Shen Yan, Tao Zhu, Zirui Wang, Yuan Cao, Mi~Zhang, Soham Ghosh, Yonghui Wu, and Jiahui Yu.
\newblock Videococa: Video-text modeling with zero-shot transfer from contrastive captioners.
\newblock {\em arXiv preprint arXiv:2212.04979}, 2022.

\bibitem{zhao2024videoprism}
Long Zhao, Nitesh~B Gundavarapu, Liangzhe Yuan, Hao Zhou, Shen Yan, Jennifer~J Sun, Luke Friedman, Rui Qian, Tobias Weyand, Yue Zhao, et~al.
\newblock Videoprism: A foundational visual encoder for video understanding.
\newblock In {\em ICML}, 2024.

\bibitem{jin2024mv}
Xiaojie Jin, Bowen Zhang, Weibo Gong, Kai Xu, Xueqing Deng, Peng Wang, Zhao Zhang, Xiaohui Shen, and Jiashi Feng.
\newblock Mv-adapter: Multimodal video transfer learning for video text retrieval.
\newblock In {\em CVPR}, 2024.

\bibitem{wu2023cap4video}
Wenhao Wu, Haipeng Luo, Bo~Fang, Jingdong Wang, and Wanli Ouyang.
\newblock Cap4video: What can auxiliary captions do for text-video retrieval?
\newblock In {\em CVPR}, 2023.

\bibitem{wang2022image}
Wenhui Wang, Hangbo Bao, Li~Dong, Johan Bjorck, Zhiliang Peng, Qiang Liu, Kriti Aggarwal, Owais~Khan Mohammed, Saksham Singhal, Subhojit Som, et~al.
\newblock Image as a foreign language: Beit pretraining for all vision and vision-language tasks.
\newblock {\em arXiv preprint arXiv:2208.10442}, 2022.

\bibitem{li2021align}
Junnan Li, Ramprasaath Selvaraju, Akhilesh Gotmare, Shafiq Joty, Caiming Xiong, and Steven Chu~Hong Hoi.
\newblock Align before fuse: Vision and language representation learning with momentum distillation.
\newblock In {\em NeurIPS}, 2021.

\bibitem{li2022blip}
Junnan Li, Dongxu Li, Caiming Xiong, and Steven Hoi.
\newblock Blip: Bootstrapping language-image pre-training for unified vision-language understanding and generation.
\newblock In {\em ICML}, 2022.

\bibitem{achiam2023gpt}
Josh Achiam, Steven Adler, Sandhini Agarwal, Lama Ahmad, Ilge Akkaya, Florencia~Leoni Aleman, Diogo Almeida, Janko Altenschmidt, Sam Altman, Shyamal Anadkat, et~al.
\newblock Gpt-4 technical report.
\newblock {\em arXiv preprint arXiv:2303.08774}, 2023.

\bibitem{lin2024vila}
Ji~Lin, Hongxu Yin, Wei Ping, Pavlo Molchanov, Mohammad Shoeybi, and Song Han.
\newblock Vila: On pre-training for visual language models.
\newblock In {\em CVPR}, 2024.

\bibitem{zhang2024internlm}
Pan Zhang, Xiaoyi Dong, Yuhang Zang, Yuhang Cao, Rui Qian, Lin Chen, Qipeng Guo, Haodong Duan, Bin Wang, Linke Ouyang, et~al.
\newblock Internlm-xcomposer-2.5: A versatile large vision language model supporting long-contextual input and output.
\newblock {\em arXiv preprint arXiv:2407.03320}, 2024.

\bibitem{chen2024far}
Zhe Chen, Weiyun Wang, Hao Tian, Shenglong Ye, Zhangwei Gao, Erfei Cui, Wenwen Tong, Kongzhi Hu, Jiapeng Luo, Zheng Ma, et~al.
\newblock How far are we to gpt-4v? closing the gap to commercial multimodal models with open-source suites.
\newblock {\em arXiv preprint arXiv:2404.16821}, 2024.

\bibitem{fu2023mme}
Chaoyou Fu, Peixian Chen, Yunhang Shen, Yulei Qin, Mengdan Zhang, Xu~Lin, Jinrui Yang, Xiawu Zheng, Ke~Li, Xing Sun, et~al.
\newblock Mme: A comprehensive evaluation benchmark for multimodal large language models.
\newblock {\em arXiv preprint arXiv:2306.13394}, 2023.

\bibitem{liu2023mmbench}
Yuan Liu, Haodong Duan, Yuanhan Zhang, Bo~Li, Songyang Zhang, Wangbo Zhao, Yike Yuan, Jiaqi Wang, Conghui He, Ziwei Liu, et~al.
\newblock Mmbench: Is your multi-modal model an all-around player?
\newblock {\em arXiv preprint arXiv:2307.06281}, 2023.

\bibitem{li2023seed}
Bohao Li, Rui Wang, Guangzhi Wang, Yuying Ge, Yixiao Ge, and Ying Shan.
\newblock Seed-bench: Benchmarking multimodal llms with generative comprehension.
\newblock {\em arXiv preprint arXiv:2307.16125}, 2023.

\bibitem{fu2024video}
Chaoyou Fu, Yuhan Dai, Yondong Luo, Lei Li, Shuhuai Ren, Renrui Zhang, Zihan Wang, Chenyu Zhou, Yunhang Shen, Mengdan Zhang, et~al.
\newblock Video-mme: The first-ever comprehensive evaluation benchmark of multi-modal llms in video analysis.
\newblock {\em arXiv preprint arXiv:2405.21075}, 2024.

\bibitem{zhou2024mlvu}
Junjie Zhou, Yan Shu, Bo~Zhao, Boya Wu, Shitao Xiao, Xi~Yang, Yongping Xiong, Bo~Zhang, Tiejun Huang, and Zheng Liu.
\newblock Mlvu: A comprehensive benchmark for multi-task long video understanding.
\newblock {\em arXiv preprint arXiv:2406.04264}, 2024.

\bibitem{xiao2021next}
Junbin Xiao, Xindi Shang, Angela Yao, and Tat-Seng Chua.
\newblock Next-qa: Next phase of question-answering to explaining temporal actions.
\newblock In {\em CVPR}, 2021.

\bibitem{gorti2022x}
Satya~Krishna Gorti, No{\"e}l Vouitsis, Junwei Ma, Keyvan Golestan, Maksims Volkovs, Animesh Garg, and Guangwei Yu.
\newblock X-pool: Cross-modal language-video attention for text-video retrieval.
\newblock In {\em CVPR}, 2022.

\bibitem{liu2022ts2}
Yuqi Liu, Pengfei Xiong, Luhui Xu, Shengming Cao, and Qin Jin.
\newblock Ts2-net: Token shift and selection transformer for text-video retrieval.
\newblock In {\em ECCV}, 2022.

\bibitem{lin2023revisiting}
Zhiqiu Lin, Xinyue Chen, Deepak Pathak, Pengchuan Zhang, and Deva Ramanan.
\newblock Revisiting the role of language priors in vision-language models.
\newblock In {\em ICML}, 2024.

\bibitem{lin2024evaluating}
Zhiqiu Lin, Deepak Pathak, Baiqi Li, Jiayao Li, Xide Xia, Graham Neubig, Pengchuan Zhang, and Deva Ramanan.
\newblock Evaluating text-to-visual generation with image-to-text generation.
\newblock In {\em ECCV}, 2024.

\bibitem{bain2021frozen}
Max Bain, Arsha Nagrani, G{\"u}l Varol, and Andrew Zisserman.
\newblock Frozen in time: A joint video and image encoder for end-to-end retrieval.
\newblock In {\em ICCV}, 2021.

\bibitem{lei2021less}
Jie Lei, Linjie Li, Luowei Zhou, Zhe Gan, Tamara~L Berg, Mohit Bansal, and Jingjing Liu.
\newblock Less is more: Clipbert for video-and-language learning via sparse sampling.
\newblock In {\em CVPR}, 2021.

\bibitem{cheng2023vindlu}
Feng Cheng, Xizi Wang, Jie Lei, David Crandall, Mohit Bansal, and Gedas Bertasius.
\newblock Vindlu: A recipe for effective video-and-language pretraining.
\newblock In {\em CVPR}, 2023.

\bibitem{zhang2018cross}
Bowen Zhang, Hexiang Hu, and Fei Sha.
\newblock Cross-modal and hierarchical modeling of video and text.
\newblock In {\em ECCV}, 2018.

\bibitem{gabeur2020multi}
Valentin Gabeur, Chen Sun, Karteek Alahari, and Cordelia Schmid.
\newblock Multi-modal transformer for video retrieval.
\newblock In {\em ECCV}, 2020.

\bibitem{yu2018joint}
Youngjae Yu, Jongseok Kim, and Gunhee Kim.
\newblock A joint sequence fusion model for video question answering and retrieval.
\newblock In {\em ECCV}, 2018.

\bibitem{miech2019howto100m}
Antoine Miech, Dimitri Zhukov, Jean-Baptiste Alayrac, Makarand Tapaswi, Ivan Laptev, and Josef Sivic.
\newblock Howto100m: Learning a text-video embedding by watching hundred million narrated video clips.
\newblock In {\em ICCV}, 2019.

\bibitem{dao2023flashattention2}
Tri Dao.
\newblock Flash{A}ttention-2: Faster attention with better parallelism and work partitioning.
\newblock In {\em ICLR}, 2024.

\bibitem{agrawal2019nocaps}
Harsh Agrawal, Karan Desai, Yufei Wang, Xinlei Chen, Rishabh Jain, Mark Johnson, Dhruv Batra, Devi Parikh, Stefan Lee, and Peter Anderson.
\newblock Nocaps: Novel object captioning at scale.
\newblock In {\em ICCV}, 2019.

\bibitem{liu2024visual}
Haotian Liu, Chunyuan Li, Qingyang Wu, and Yong~Jae Lee.
\newblock Visual instruction tuning.
\newblock In {\em NeurIPS}, 2023.

\bibitem{zhou2018towards}
Luowei Zhou, Chenliang Xu, and Jason Corso.
\newblock Towards automatic learning of procedures from web instructional videos.
\newblock In {\em AAAI}, 2018.

\bibitem{zhang2024lmms}
Kaichen Zhang, Bo~Li, Peiyuan Zhang, Fanyi Pu, Joshua~Adrian Cahyono, Kairui Hu, Shuai Liu, Yuanhan Zhang, Jingkang Yang, Chunyuan Li, et~al.
\newblock Lmms-eval: Reality check on the evaluation of large multimodal models.
\newblock {\em arXiv preprint arXiv:2407.12772}, 2024.

\bibitem{cai2024temporalbench}
Mu~Cai, Reuben Tan, Jianrui Zhang, Bocheng Zou, Kai Zhang, Feng Yao, Fangrui Zhu, Jing Gu, Yiwu Zhong, Yuzhang Shang, et~al.
\newblock Temporalbench: Benchmarking fine-grained temporal understanding for multimodal video models.
\newblock {\em arXiv preprint arXiv:2410.10818}, 2024.

\bibitem{cohen2009pearson}
Israel Cohen, Yiteng Huang, Jingdong Chen, Jacob Benesty, Jacob Benesty, Jingdong Chen, Yiteng Huang, and Israel Cohen.
\newblock Pearson correlation coefficient.
\newblock {\em Noise reduction in speech processing}, 2009.

\bibitem{chen2011collecting}
David Chen and William~B Dolan.
\newblock Collecting highly parallel data for paraphrase evaluation.
\newblock In {\em ACL}, 2011.

\bibitem{wang2019vatex}
Xin Wang, Jiawei Wu, Junkun Chen, Lei Li, Yuan-Fang Wang, and William~Yang Wang.
\newblock Vatex: A large-scale, high-quality multilingual dataset for video-and-language research.
\newblock In {\em ICCV}, 2019.

\bibitem{chen2024rextime}
Jr-Jen Chen, Yu-Chien Liao, Hsi-Che Lin, Yu-Chu Yu, Yen-Chun Chen, and Yu-Chiang~Frank Wang.
\newblock Rextime: A benchmark suite for reasoning-across-time in videos.
\newblock In {\em NeurIPS}, 2024.

\end{thebibliography}
}

\newpage 
\appendix
\section*{\Large{Appendix}}

The appendix is organized into the following sections:
\begin{itemize}
    \item Appendix~\ref{sup:dataset}: Dataset Details
        \begin{itemize}
            \item \ref{sup:dataset_tvr} Text-Video Retrieval
            \item \ref{sup:dataset_benchmark} Comprehensive Multi-Modal  Understanding
        \end{itemize}
    \item Appendix~\ref{sup:implementation}: Implementation Details
    \item Appendix~\ref{sup:inference}: Inference Details of BLiM
    \item Appendix~\ref{sup:proof}: Proof of Proposition 1    
    \item Appendix~\ref{sup:cpn}: Further Discussion on CPN
        \begin{itemize}
            \item \ref{sup:cpn_bias} Alleviation of Candidate Prior Bias
            \item \ref{sup:cpn_captioning} CPN Decoding in Visual Captioning
            \item \ref{sup:text_prior} Analysis on Text Candidate Prior
            \item \ref{sup:computational} Discussion on Computational Cost
        \end{itemize}
    \item Appendix~\ref{sup:quantitative}: Further Quantitative Results
        \begin{itemize}
            \item \ref{sup:multi_text} Results on Multi-Text Retrieval Settings
            \item \ref{sup:alpha} Sensitivity Study of $\alpha$ in CPN
            \item \ref{sup:ble_full} Results on Bidirectional Likelihood Estimation
            \item \ref{sup:cpn_full} Results on Candidate Prior Normalization
        \end{itemize}
    \item Appendix~\ref{sup:qualitative}: Further Qualitative Results
        \begin{itemize}
            \item \ref{sup:ble_qual} Results on Bidirectional Likelihood Estimation
            \item \ref{sup:cpn_qual} Results on Candidate Prior Normalization
            \item \ref{sup:instruction} Results on Instruction-based Retrieval
        \end{itemize}
\end{itemize}

\section{Dataset Details}
\label{sup:dataset}

\subsection{Text-Video Retrieval}
\label{sup:dataset_tvr}

\noindent \textbf{DiDeMo~\citep{anne2017localizing}.} 
Distinct Describable Moments (DiDeMo) contains 10K videos which are divided into 5-second segments. 
It has a total of 26K moments whose descriptions are detailed and contain camera movement, temporal transition indicators, and activities.
We follow the previous works~\cite{bain2021frozen,lei2021less,luo2022clip4clip,wu2023cap4video,cheng2023vindlu,li2023unmasked} by concatenating all captions of one video and solving the task as a paragraph-video retrieval task.
The number of training and test samples is 8,394 and 1,003, respectively.

\noindent \textbf{ActivityNet~\citep{caba2015activitynet}.} 
ActivityNet dataset contains 19K videos from YouTube, which are categorized into 200 different types of activities. 
On average, each category has 137 videos and each video has 1.41 activities which are annotated with temporal boundaries.
Similar to DiDeMo, we also concatenate all the captions of a video to form a paragraph-video retrieval task on the `val1' split by following \cite{zhang2018cross,gabeur2020multi,luo2022clip4clip,cheng2023vindlu,li2023unmasked}.
Therefore, the number of training and test samples is 10,009 and 4,917, respectively.

\noindent \textbf{LSMDC~\citep{rohrbach2017movie}.} 
Large Scale Movie Description Challenge (LSMDC) contains 118K short video clips from 202 movies with captions from the movie script or from transcribed DVS (descriptive video services) for the visually impaired. 
Our model is trained with 101,055 videos and evaluated on 1,000 videos.

\noindent \textbf{MSRVTT~\citep{xu2016msr}.} 
Microsoft Research Video to Text (MSRVTT) contains 10K video clips from 20 categories, with each video clip annotated with 20 sentences. There are 29K unique words in all captions.
Following the literature~\cite{yu2018joint,luo2022clip4clip,miech2019howto100m,gabeur2020multi,wu2023cap4video,cheng2023vindlu,li2023unmasked}, we train our model with 9,000 $\times$ 20 training samples and 1,000 test samples.

\subsection{Comprehensive Multi-Modal Understanding}
\label{sup:dataset_benchmark}

\noindent \textbf{MME~\citep{fu2023mme}.} 
Multi-modal large language Model Evaluation benchmark (MME) is composed of 14 subtasks where all the samples are manually annotated.
MME targets to assess MLLMs' perception and cognition abilities including OCR, existence of objects, commonsense reasoning, numerical calculation, code reasoning, etc.

\noindent \textbf{MMBench~\citep{liu2023mmbench}.}
MMBench is a bilingual benchmark to evaluate the MLLMs' multi-modal understanding abilities.
This benchmark includes multiple-choice questions across the 20 ability dimensions like spatial relationship, physical property, attribute recognition, object localization, etc.

\noindent \textbf{SeedBench~\citep{li2023seed}.}
SeedBench aims at a comprehensive assessment of generative models and contains 19K manually annotated multiple-choice questions across the 12 ability dimensions both on the image and video domain.
The questions cover both spatial and temporal understanding like scene understanding, action prediction, procedure understanding, etc.

\noindent \textbf{MVBench~\citep{li2023mvbench}.}
Multi-modal Video understanding Benchmark (MVBench) consists of 20 challenging video understanding tasks that can effectively assess the ability to comprehend temporal evolution in dynamic videos. 
It consists of 9 main tasks for spatial understanding, which are then further split into a total of 20 tasks for temporal understanding. 

\noindent \textbf{VideoMME~\citep{fu2024video}.}
Multi-Modal Evaluation benchmark of MLLMs in Video analysis (VideoMME) evaluates the ability of MLLMs to handle sequential visual data on 6 primary visual domains with 30 subcategories.
The videos are categorized as short, medium, and long, ranging from 11 seconds to 1 hour.
A total of 900 videos are in the benchmark with 2,700 questions.

\noindent \textbf{MLVU~\citep{zhou2024mlvu}.}
Multi-task Long Video Understanding benchmark (MLVU) targets to assess long video understanding performance spanning 7 video genres including movies, egocentric videos, cartoons, etc.
MLVU contains 2,593 questions on 9 categories like topic reasoning, plot question answering, action count, ego reasoning, etc.

\noindent \textbf{NExT-QA~\citep{xiao2021next}.}
NExT-QA is a video question answering task aiming to evaluate causal action reasoning, temporal action reasoning, and common scene comprehension.
This dataset includes 47,692 multiple-choice questions and 52,044 open-ended questions on a total of 5,440 videos.
\begin{figure*}[!t]
    \centering
    \begin{subfigure}[h]{0.49\linewidth}
        \includegraphics[width=1.0\linewidth]{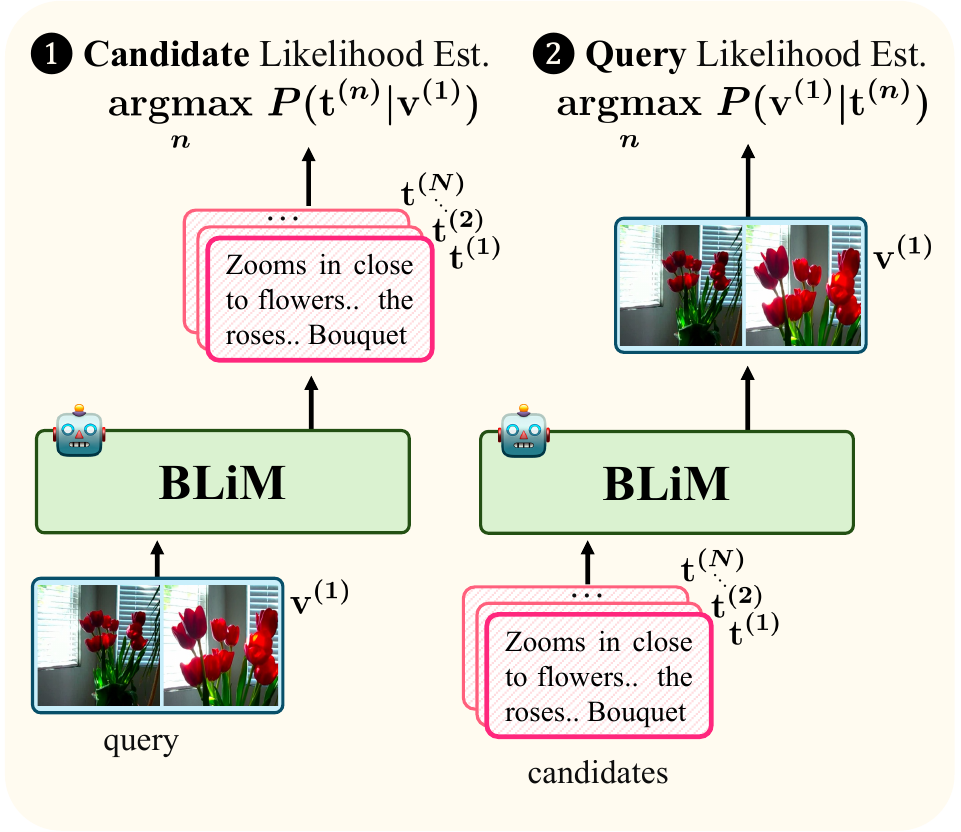}
        \caption{\textbf{Video-to-Text Retrieval.}}
        \label{fig:inference1}
    \end{subfigure}
    \begin{subfigure}[h]{0.49\linewidth}
        \includegraphics[width=1.0\linewidth]{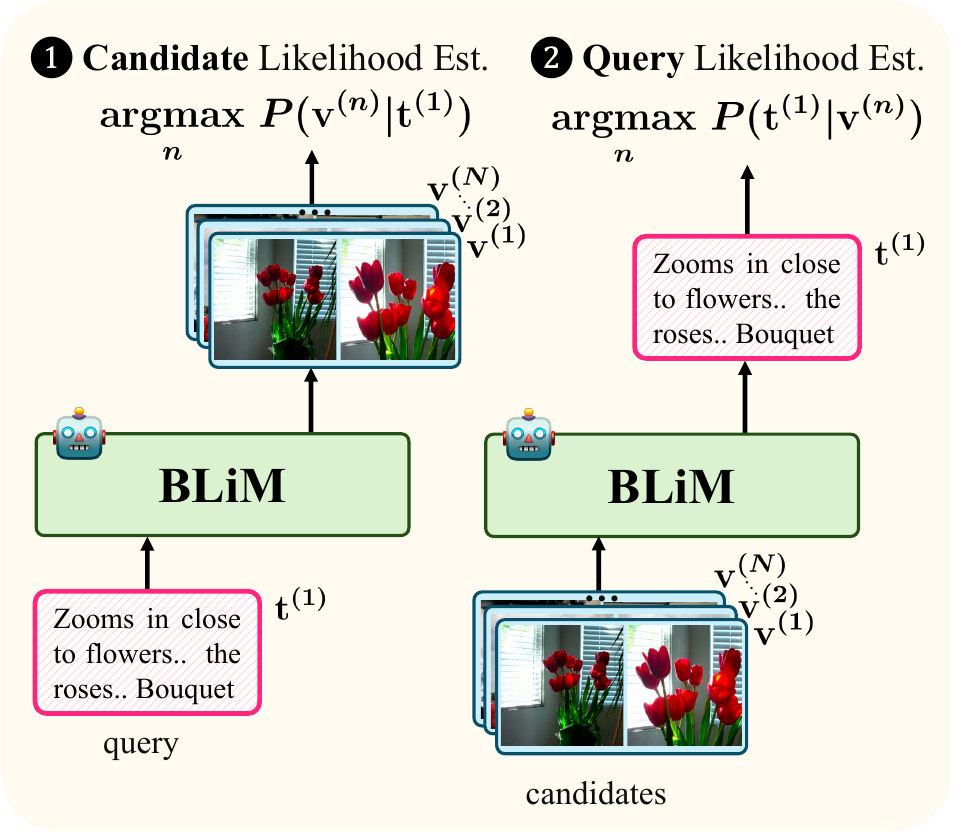}
        \caption{\textbf{Text-to-Video Retrieval.}}
        \label{fig:inference2}
    \end{subfigure}
    \caption{Inference details of BLiM in (a) video-to-text and (b) text-to-video retrievals.}
    \label{fig:inference}
\end{figure*}
\section{Implementation Details}
\label{sup:implementation}

\noindent \textbf{BLiM details.}
Our BLiM is built upon VideoChat-Flash~\citep{li2023mvbench} and is further fine-tuned on each Text-Video Retrieval dataset.
Specifically, VideoChat-Flash consists of a video encoder, a linear projection layer, and a LLM.
The visual encoder and LLM are initialized with UMT-L~\citep{li2023unmasked} and Qwen2~\citep{yang2024qwen2}, respectively.
We freeze parameters in the video encoder and LLM, and only update parameters in the linear projection layer and LoRA for parameter-efficient fine-tuning, resulting in 10M trainable parameters among 7B total parameters (8\%).
We accumulate gradients from $P(\mathbf{t}|\mathbf{v})$ and $P(\mathbf{v}|\mathbf{t})$, and update the trainable parameters at once.

\begin{table}[!t]
    \centering
    \begin{adjustbox}{width=\linewidth}
    \begin{tabular}{l|c c c c}
        \toprule
        & DiDeMo & ActivityNet & LSMDC & MSRVTT \\
        \midrule
        \midrule
        optimizer & \multicolumn{4}{c}{AdamW} \\ 
        optimizer momentum & \multicolumn{4}{c}{$\beta_1 = 0.9$, $\beta_2 = 0.95$} \\
        weight decay & \multicolumn{4}{c}{1.0} \\
        warmup epochs & \multicolumn{4}{c}{1} \\
        input frames & \multicolumn{4}{c}{16} \\
        \midrule
        $\alpha$ for $P^\alpha(\mathbf{t}|\mathbf{v})$ & 0.8 & 0.9 & 1.0 & 0.9 \\
        $\alpha$ for $P^\alpha(\mathbf{v}|\mathbf{t})$ & 0.0 & 0.2 & 0.2 & 0.0 \\
        total epochs & 5 & 5 & 3 & 3 \\
        learning rate & 2e-4 & 1e-4 & 1e-4 & 1e-4 \\
        batch size & 32 & 32 & 256 & 512 \\
        \bottomrule
    \end{tabular}
    \end{adjustbox}
    \caption{\textbf{Experimental settings in Text-Video Retrieval.}}
    \label{tab:detail}
\end{table}
\noindent \textbf{Experimental settings.}
The self-attention mechanism in our model is implemented under FlashAttention2~\citep{dao2023flashattention2} and we sample 16 frames per video for all datasets.
These 16 frames are divided into four clips with four frames each.
The learning rate is 2e-4 for DiDeMo and 1e-4 for ActivityNet, LSMDC, and MSRVTT with AdamW optimizer.
We train our model on 8 $\times$ A6000 GPUs with a batch size of 32, 32, 256, and 512 for DiDeMo, ActivityNet, LSMDC, and MSRVTT, respectively.
For inference, we select the top-16 candidates according to the similarity from InternVideo2 1B~\cite{wang2024internvideo2} and rerank them by leveraging bidirectional likelihoods.
More details are summarized in Tab.~\ref{tab:detail}.
\section{Inference Details of BLiM}
\label{sup:inference}

In inference, BLiM calculates candidate and query likelihood, and ensembles them for final prediction.
Fig.~\ref{fig:inference1} and \ref{fig:inference2} illustrate the inference procedure of video-to-text and text-to-video retrieval, respectively.
For example, on candidate likelihood estimation in Fig.~\ref{fig:inference1} (left) and \ref{fig:inference2} (left), we fix the \textit{input} of the model as a video (or text) query and seek the best text (or video) content by replacing the \textit{output} with text (or video) candidates.
On the other hand, on query likelihood estimation in Fig.~\ref{fig:inference1} (right) and \ref{fig:inference2} (right), we fix the \textit{output} of the model as a text (or video) query and seek the best video (or text) content by replacing the \textit{input} with video (or text) candidates.
\begin{figure*}[!t] 
    \centering
    \begin{subfigure}[h]{0.24\linewidth}
        \includegraphics[width=1.0\linewidth]{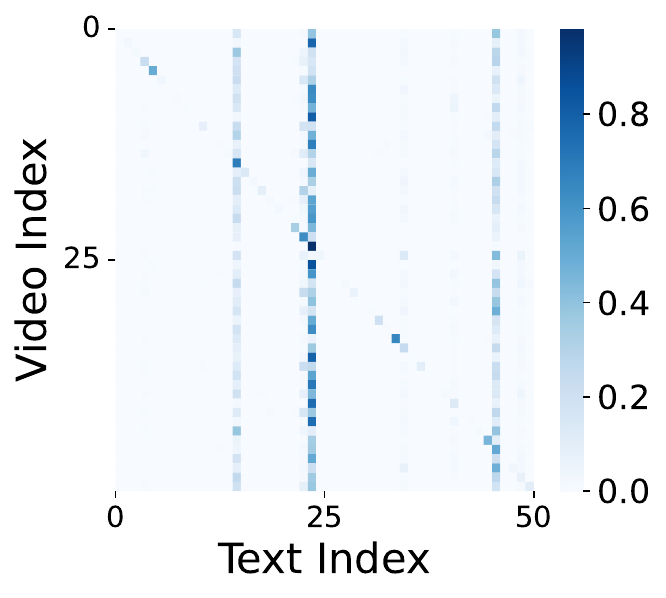}
        \caption{w/o CPN in video-to-text.}
        \label{fig:heatmap1}
    \end{subfigure}
    \begin{subfigure}[h]{0.24\linewidth}
        \includegraphics[width=1.0\linewidth]{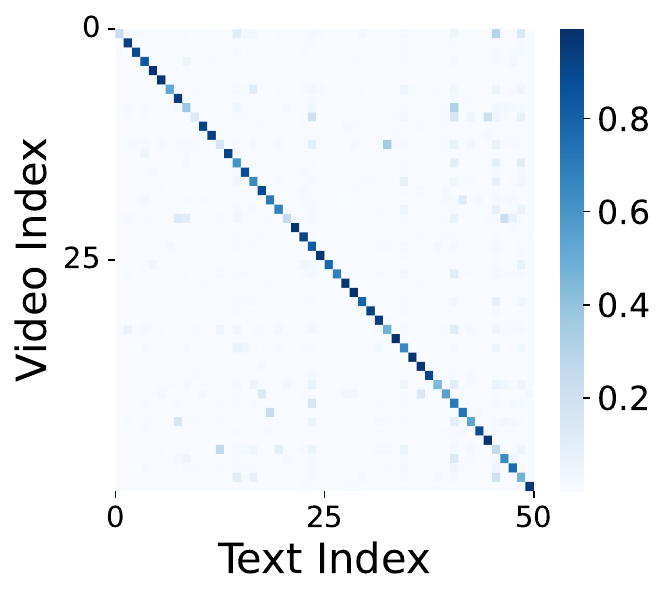}
        \caption{w/ CPN in video-to-text.}
        \label{fig:heatmap2}
    \end{subfigure}
    \begin{subfigure}[h]{0.24\linewidth}
        \includegraphics[width=1.0\linewidth]{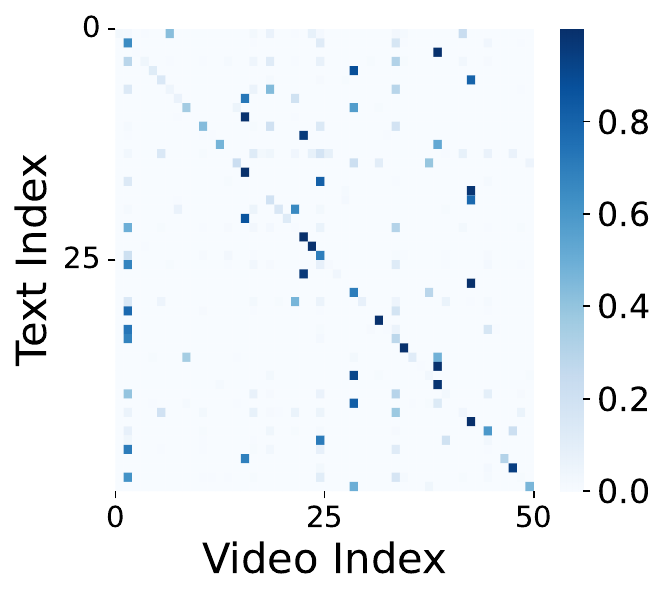}
        \caption{w/o CPN in text-to-video.}
        \label{fig:heatmap3}
    \end{subfigure}
    \begin{subfigure}[h]{0.24\linewidth}
        \includegraphics[width=1.0\linewidth]{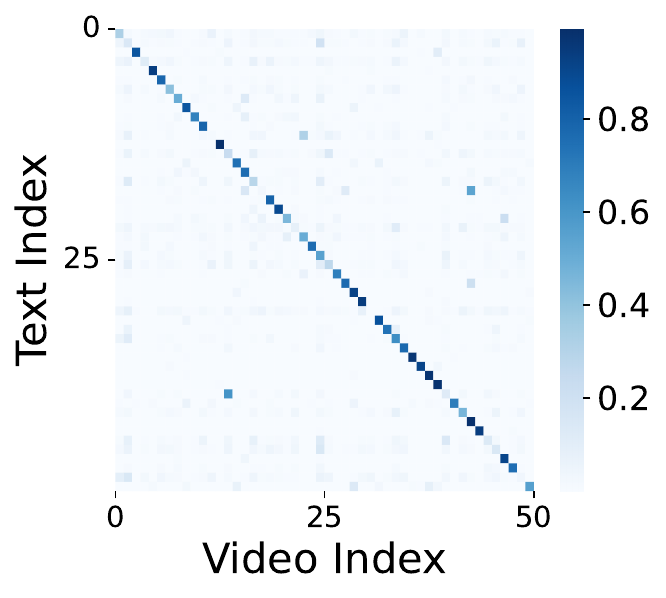}
        \caption{w/ CPN in text-to-video.}
        \label{fig:heatmap4}
    \end{subfigure}
    \caption{\textbf{Visualization of retrieval results on the candidate likelihood estimation w/ and w/o CPN.}
    50 text-video pairs are sampled to avoid visual clutter.
    }
    \label{fig:heatmap}
\end{figure*}
\section{Proof of Proposition 1}
\label{sup:proof}

\noindent \textbf{Proposition 1.}
\textit{Let $P(\mathbf{t}^{(m)}|\mathbf{v}^{(m)})$ denote the candidate likelihood for retrieving the most relevant text $\mathbf{t}^{(m)}$ given a query video $\mathbf{v}^{(m)}$. Suppose that:}
\begin{enumerate}
    \item \textit{The query likelihood correctly ranks $\mathbf{t}^{(m)}$ over any negative sample $\mathbf{t}^{(n)}$ and the gap is bounded as:}
    \begin{align}
        0 < \log P(\mathbf{v}^{(m)}|\mathbf{t}^{(m)}) - \log P(\mathbf{v}^{(m)}|\mathbf{t}^{(n)}) < \varepsilon.
        \label{eq:query_gap2}
    \end{align}
    \item \textit{There exists a text candidate $\mathbf{t}^{(n)}$ with a larger prior probability gap:}
    \begin{align}
        \log P(\mathbf{t}^{(n)}) - \log P(\mathbf{t}^{(m)}) > c\varepsilon, \;\; \text{for some } c > 1.
        \label{eq:prior_gap2}
    \end{align}
\end{enumerate}
\textit{Then, the candidate likelihood ranking is reversed:}
\begin{align}
    P(\mathbf{t}^{(m)}|\mathbf{v}^{(m)}) < P(\mathbf{t}^{(n)}|\mathbf{v}^{(m)}).
\end{align}

\begin{proof}
    The candidate likelihood cap between $\mathbf{t}^{(m)}$ and $\mathbf{t}^{(n)}$ given the video query $\mathbf{v}^{(m)}$ is written as:
    \begin{align}
        & \log P(\mathbf{t}^{(m)}|\mathbf{v}^{(m)}) - \log P(\mathbf{t}^{(n)}|\mathbf{v}^{(m)}) \\ 
        & = \log P(\mathbf{v}^{(m)}|\mathbf{t}^{(m)}) + \log P(\mathbf{t}^{(m)}) \nonumber \\
        & \quad - \log P(\mathbf{v}^{(m)}|\mathbf{t}^{(n)}) - \log P(\mathbf{t}^{(n)}) && \text{(by Bayes' Rule)} \\
        & < \varepsilon + \log P(\mathbf{t}^{(m)}) - \log P(\mathbf{t}^{(n)}) && \text{(by Eq.~\eqref{eq:query_gap2})} \\
         & < \varepsilon - c\varepsilon = \varepsilon (1-c) && \text{(by Eq.~\eqref{eq:prior_gap2})} \\
         & < 0. \quad \text{(by \( c > 1 \))}
    \end{align}
    Therefore, $P(\mathbf{t}^{(m)} |\mathbf{v}^{(m)}) < P(\mathbf{t}^{(n)}|\mathbf{v}^{(m)})$.
\end{proof}
This proposition indicates that the candidate likelihood ranking is reversed, leading to the retrieval of an incorrect candidate, although the query likelihood identifies the accurate candidate in Eq.~\eqref{eq:query_gap2}.
The inaccurate relevance prediction arises due to a substantial gap in candidate prior probabilities, as shown in Eq.~\eqref{eq:prior_gap2}.
This motivates us to jointly consider query and candidate likelihood (\ie, Bidirectional Likelihood Estimation) along with CPN to mitigate bias towards candidate prior probability.
\section{Further Discussion on CPN}
\label{sup:cpn}

\subsection{Alleviation of Candidate Prior Bias}
\label{sup:cpn_bias}

To verify the alleviation of candidate prior bias, we provide heatmaps in Fig.~\ref{fig:heatmap} w/ and w/o CPN on the candidate likelihood estimation.
For example, in video-to-text retrieval, the candidate likelihood estimation w/o CPN demonstrates suboptimal retrieval results since the text with the highest prior probability, \ie, the 24th text, is retrieved for most videos.
On the other hand, the candidate likelihood w/ CPN leads to a balanced prediction where each text is retrieved for its own paired video in Fig.~\ref{fig:heatmap2}.
This reveals that CPN successfully alleviates candidate prior bias and encourages the model to consider text-video correspondences more.
Furthermore, candidate prior bias is more pronounced in video-to-text retrieval due to the high reliance of MLLMs on LLMs' pretrained knowledge.
This becomes evident when comparing Fig.~\ref{fig:heatmap1} and Fig.~\ref{fig:heatmap3}, a clear vertical line is observed on video-to-text retrieval in Fig.~\ref{fig:heatmap1}.

\subsection{CPN Decoding in Visual Captioning}
\label{sup:cpn_captioning}

\begin{table}[!t]
    \centering
    \begin{adjustbox}{width=\linewidth}
    \begin{tabular}{l|c|c|c|c|c|c}
        \toprule
        & \multicolumn{1}{c|}{\textbf{COCO}} & \multicolumn{1}{c|}{\textbf{NoCaps}} & \multicolumn{1}{c|}{\textbf{LLaVA-Wild}} & \multicolumn{1}{c|}{\textbf{YouCook2}} & \multicolumn{1}{c|}{\textbf{VDC}} & \multicolumn{1}{c}{\textbf{TemporalBench}} \\
        \midrule
        \midrule
        LLaVA-Onevision~\cite{li2024llava} & 140.5 & 87.7 & 83.2 & 19.0 & 2.5 & 36.1 \\
        \rowcolor[HTML]{BFF2FF}
        LLaVA-Onevision$^\dagger$ (\textbf{Ours}) & \textbf{142.1} & \textbf{89.9} & \textbf{84.1} & \textbf{22.4} & \textbf{3.0} & \textbf{37.6} \\
        \bottomrule
    \end{tabular}
    \end{adjustbox}
    \caption{\textbf{Results on visual captioning.}
    We report CIDEr for COCO, NoCaps, and YouCook2, and average GPT score for LLaVA-Wild and VideoDetailCaption (VDC).
    The TemporalBench score is reported for TemporalBench, which is based on the embedding similarity.
    }
    \label{tab:captioning}
\end{table}

\begin{figure*}[!ht] 
    \centering
    \begin{subfigure}[h]{0.49\linewidth}
        \includegraphics[width=1.0\linewidth]{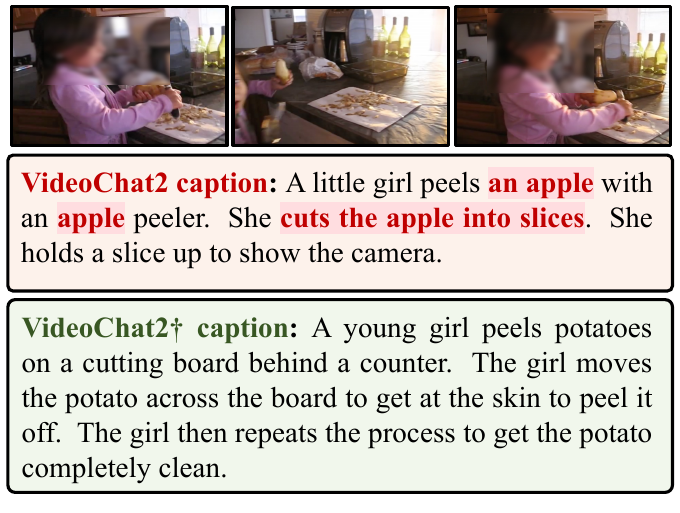}
        \caption{}
        \label{fig:caption1}
    \end{subfigure}
    \begin{subfigure}[h]{0.49\linewidth}
        \includegraphics[width=1.0\linewidth]{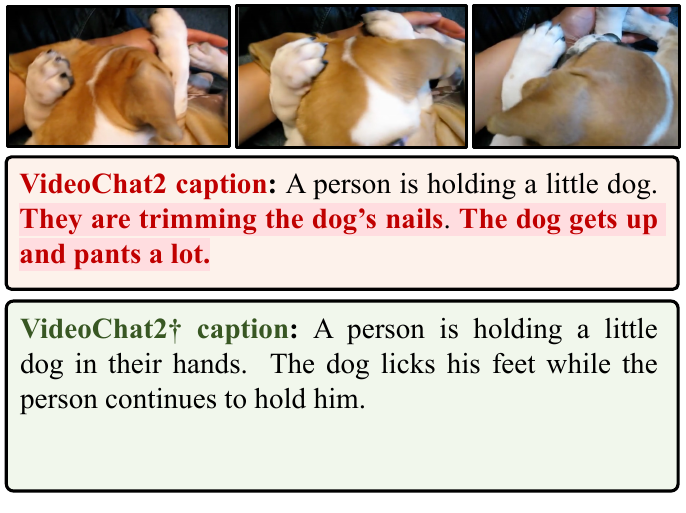}
        \caption{}
        \label{fig:caption2}
    \end{subfigure}
    \caption{\textbf{Qualitative results of CPN decoding in video captioning on ActivityNet.}
    $\dagger$ stands for the model with CPN decoding.
    The hallucinated text is highlighted in red.
    }
    \label{fig:caption}
\end{figure*}

\begin{table*}[!t]
    \centering
    \begin{adjustbox}{width=\textwidth}
    \begin{tabular}{l|c c c c c c c|c}
        \toprule
        \textbf{Model} & \textbf{MME} & \textbf{MMBench} & \textbf{MVBench} & \textbf{VideoMME} & \textbf{MLVU} & \textbf{NExT-QA} & \textbf{SeedBench} & \textbf{avg. $\Delta$} \\
        \midrule
        \midrule
        VideoChat2~\citep{li2023mvbench} & 1505.7 (1.5) & 63.9 (1.2) & 60.1 (2.4) & 42.2 (4.1) & 45.8 (6.9) & 78.9 (1.4) & 61.2 (0.9) & - \\
        VideoChat2$^\dagger$ (\textbf{Ours}) & \textbf{1607.0} (2.0) & \textbf{66.2} (1.2) & \textbf{62.3} (2.4) & \textbf{47.1} (4.1) & \textbf{48.5} (7.1) & \textbf{79.4} (1.5) & \textbf{61.7} (1.0) & \textbf{+16.3} (+4.9\%) \\
        \bottomrule
    \end{tabular}
    \end{adjustbox}
    \caption{\textbf{Inference time comparison of CPN decoding.}
    The inference time (seconds per sample) is reported in parentheses.
    $\dagger$ stands for the model with CPN decoding.
    }
    \label{tab:speed}
\end{table*}
Tab.~\ref{tab:captioning} demonstrates the quantitative results of CPN decoding to visual captioning.
We apply CPN decoding to LLaVA-Onevision~\cite{li2024llava} and evaluate its performance on six benchmarks (COCO~\cite{lin2014microsoft}, NoCaps~\cite{agrawal2019nocaps}, LLaVA-Wild~\cite{liu2024visual}, YouCook2~\cite{zhou2018towards}, VideoDetailCaption~\cite{zhang2024lmms}, and TemporalBench~\cite{cai2024temporalbench}) covering both image and video captioning tasks.
Our results show that CPN decoding consistently enhances performance across all datasets, underscoring its effectiveness in visual captioning.

To show how CPN decoding improves the performance in visual captioning, we provide qualitative results in Fig.~\ref{fig:caption} by applying CPN decoding to VideoChat2~\cite{li2023mvbench}.
The standard VideoChat2 usually generates a hallucinated text by overlooking the visual content.
For example, in Fig.~\ref{fig:caption1}, the word `apple' is hallucinated which does not appear in the video.
Similarly, in Fig.~\ref{fig:caption2}, the standard VideoChat2 also generates a hallucinated phrase ``They are trimming the dog's nails'' while the dog licks his feet in the video.
However, with our CPN decoding (denoted as VideoChat2$^\dagger$), the hallucinated text is successfully removed by encouraging the model to take into account visual contents more.

\subsection{Analysis on Text Candidate Prior}
\label{sup:text_prior}

\begin{figure}[!t]
    \centering
    \begin{subfigure}[h]{0.49\linewidth}
        \includegraphics[width=1.0\linewidth]{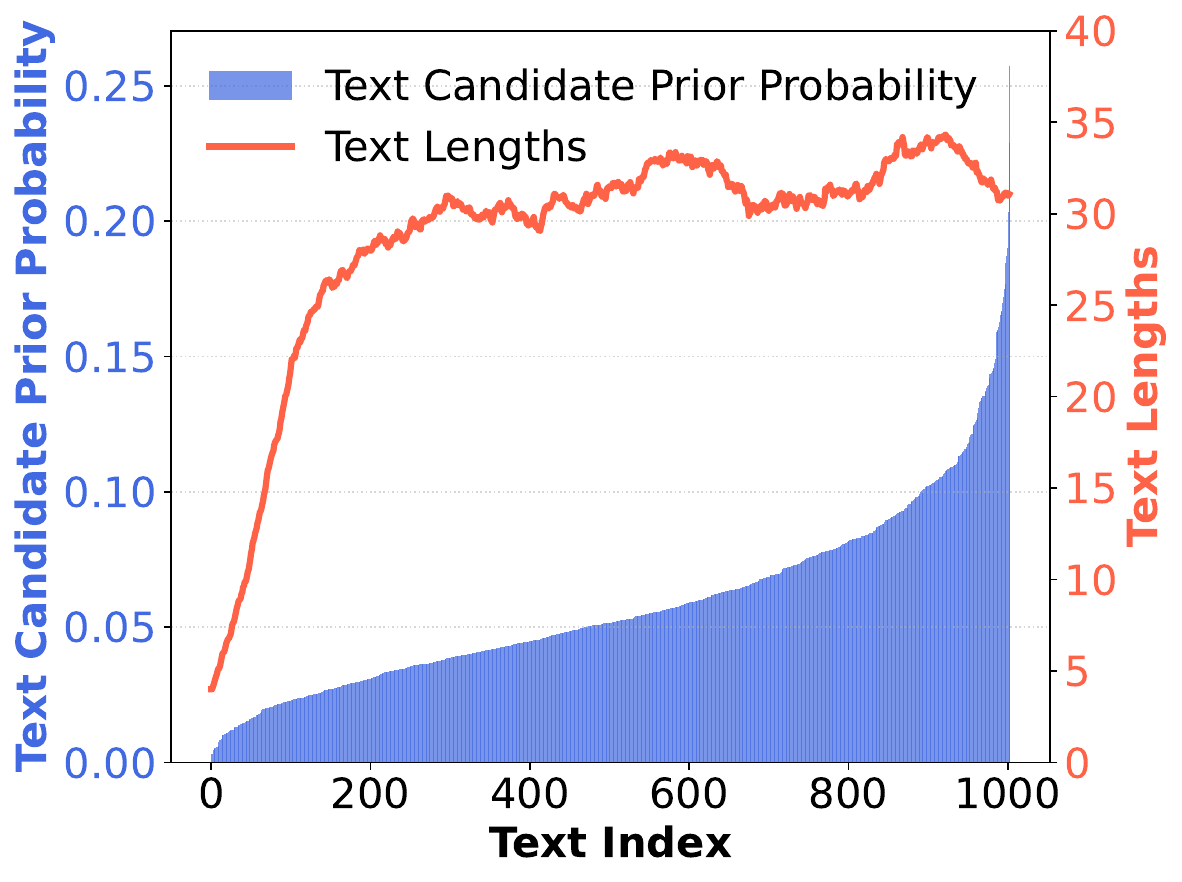}
        \caption{\textbf{Prior vs Text Lengths.}}
        \label{fig:text_prior1}
    \end{subfigure}
    \begin{subfigure}[h]{0.49\linewidth}
        \includegraphics[width=1.0\linewidth]{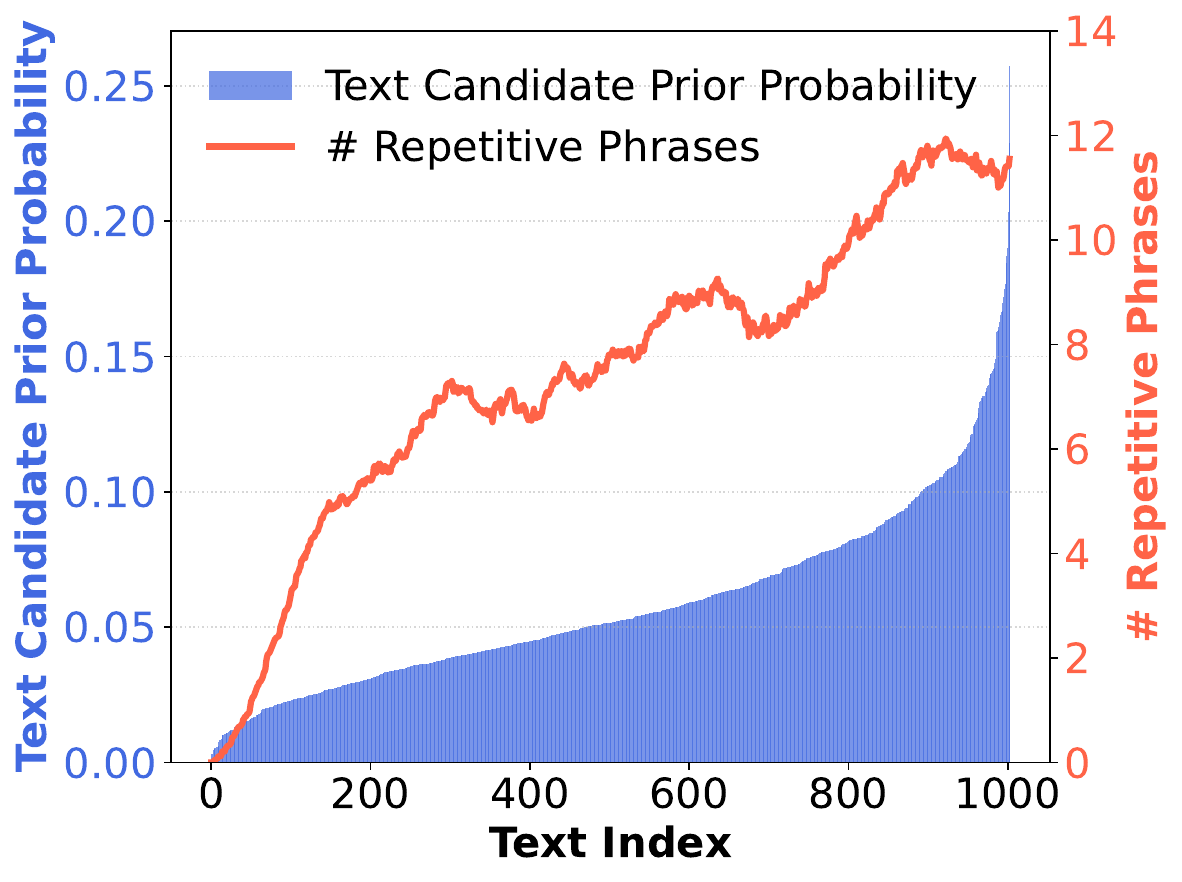}
        \caption{\textbf{Prior vs Repetitive Phrases.}}
        \label{fig:text_prior2}
    \end{subfigure}
    \caption{Visualization of the correlation between (a) prior probabilities and text length and (b) prior probabilities and the number of repetitive phrases. 
    The texts are sorted in ascending order based on prior probabilities.
    }
    \label{fig:text_prior}
\end{figure}
We visualize the correlation between text candidate prior probabilities and text lengths in Fig.~\ref{fig:text_prior1}, as well as the correlation between text candidate prior probabilities and the number of repetitive phrases in Fig.~\ref{fig:text_prior2}.
Interestingly, both text length and the number of repetitive phrases increase as the text candidate prior probability increases.
Using the Pearson Correlation Coefficient~\cite{cohen2009pearson}, we find that the correlation in Fig.~\ref{fig:text_prior1} is 0.97, and that in Fig.~\ref{fig:text_prior2} is 0.93, indicating a strong relationship between text candidate prior probabilities and these linguistic properties.

\subsection{Dicussion on Computational Cost}
\label{sup:computational}

Finally, Tab.~\ref{tab:speed} demonstrates the additional inference time overhead of CPN decoding on the benchmarks in Tab. 5 of the main paper.
Since these benchmarks consist of multi-choice questions, the number of newly generated tokens by the model is less than 10 tokens.
This implies that CPN decoding introduces only a marginal increase in inference time.
In Tab.~\ref{tab:speed}, the average performance is improved by 16.3 while the additional inference time is only increased by 4.9\%.
On the other hand, the inference time might be increased if the number of newly generated tokens becomes large.
\begin{table}[!t]
    \centering
    \begin{adjustbox}{width=\linewidth}
    \begin{tabular}{l|c|c c c|c}
        \toprule
        & & Cap4Video~\cite{wu2023cap4video} & UMT~\cite{li2023unmasked} & InternVideo2 6B~\cite{wang2024internvideo2} & \cellcolor[HTML]{BFF2FF}BLiM \\
        \midrule
        \midrule
        \multirow{2}{*}{MSVD} & T2V & 51.8 & 58.2 & 61.4 & \cellcolor[HTML]{BFF2FF}\textbf{63.2}\\
        & V2T & - & 82.4 & 85.2 & \cellcolor[HTML]{BFF2FF}\textbf{85.7} \\
        \midrule
        \multirow{2}{*}{VATEX} & T2V & 66.6 & 72.0 & 75.5 & \cellcolor[HTML]{BFF2FF}\textbf{78.2} \\
        & V2T & - & 86.0 & \textbf{89.3} & \cellcolor[HTML]{BFF2FF}83.9 \\
        \bottomrule
    \end{tabular}
    \end{adjustbox}
    \caption{\textbf{Results on multi-text Text-Video Retrieval.}
    We only report R@1 both in text-to-video (T2V) and video-to-text (V2T) retrieval.}
    \label{tab:multi}
\end{table}

\begin{figure}[!t] 
    \centering
    \includegraphics[width=1.0\linewidth]{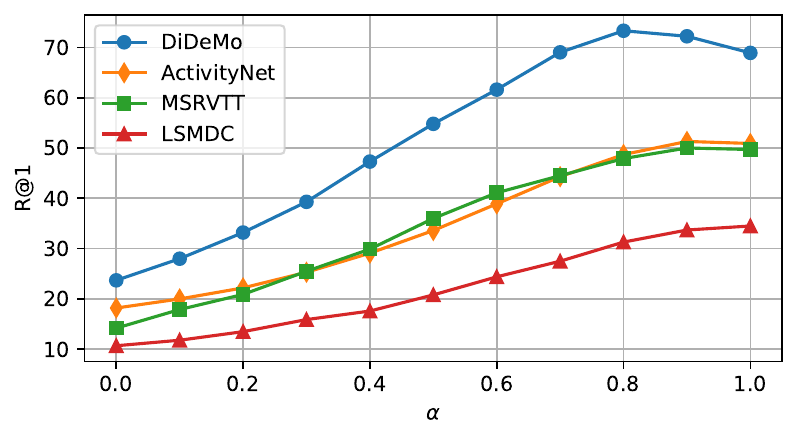}
    \caption{\textbf{Video-to-text retrieval performance on various $\alpha$.}}
    \label{fig:alpha}
\end{figure}
\section{Further Quantitative Results}
\label{sup:quantitative}

\subsection{Results on Multi-Text Retrieval Settings}
\label{sup:multi_text}

Tab.~\ref{tab:multi} demonstrates the result of BLiM in multi-text Text-Video Retrieval on MSVD~\cite{chen2011collecting} and VATEX~\cite{wang2019vatex}. 
In text-to-video retrieval on VATEX, BLiM surpasses InternVideo2 6B by 2.7. Consequently, BLiM achieves a new state-of-the-art performance in 3 out of 4 settings.

\subsection{Sensitivity Study of $\alpha$ in CPN}
\label{sup:alpha}

Fig.~\ref{fig:alpha} presents the video-to-text retrieval performance across various values of $\alpha$ in CPN (Eq. (8) of the main paper).
$\alpha = 0$ indicates that CPN is not applied to the prediction. 
Our findings reveal that an $\alpha$ range from 0.8 to 1.0 consistently yields the best performance across all datasets. 
This highlights the importance of mitigating the influence of candidate priors in candidate likelihood through the application of CPN.

\subsection{Results on Bidirectional Likelihood Estimation}
\label{sup:ble_full}

In Tab.~\ref{tab:ble_full}, we provide detailed results on bidirectional likelihood estimation.
In text-to-video retrieval, R@1 is improved by 40.1, 40.2, 26.1, and 24.3 increase on DiDeMo, ActivityNet, LSMDC, and MSRVTT, respectively.
Similarly, by reducing the effect of text candidate prior in video-to-text retrieval, a dramatic performance gain is observed in query likelihood estimation, with R@1 increasing by 36.0, 40.8, 22.8, and 35.7 on each dataset.
Finally, bidirectional likelihood estimation (BLE) further enhances performance beyond query likelihood estimation, especially in video-to-text retrieval.

\subsection{Results on Candidate Prior Normalization}
\label{sup:cpn_full}

Tab.~\ref{tab:cpn_full} demonstrates detailed results on CPN.
First, in video-to-text retrieval, we observe a substantial performance improvement after applying CPN to candidate likelihood estimation, with R@1 gains of 49.6, 33.1, 23.8, and 35.8 on each dataset.
We hypothesize that candidate prior bias is more pronounced in textual candidates, \ie, video-to-text retrieval, due to the powerful LLM's pretrained knowledge in MLLM. 
On the other hand, the performance gain is relatively marginal in text-to-video retrieval since video representations are inherently less influenced by LLM's knowledge.
Overall, incorporating CPN leads to an average R@1 improvement of 8.5 in bidirectional likelihood estimation. 
\begin{table}[!t]
    \centering
    \begin{adjustbox}{width=\linewidth}
    \begin{tabular}{l|c c|c c|c c|c c}
        \toprule
        & \multicolumn{2}{c|}{\textbf{DiDeMo}}  & \multicolumn{2}{c|}{\textbf{ActivtyNet}} & \multicolumn{2}{c|}{\textbf{LSMDC}} & \multicolumn{2}{c}{\textbf{MSRVTT}} \\
        & T2V & V2T & T2V & V2T & T2V & V2T & T2V & V2T \\
        \midrule
        \midrule
        CLE & 45.1 & 23.7 & 39.8 & 18.2 & 27.7 & 10.7 & 38.5 & 14.2 \\
        QLE & 85.2 & 59.7 & \textbf{80.0} & 59.0 & \textbf{53.8} & 33.5 & \textbf{62.8} & 49.9 \\
        \rowcolor[HTML]{BFF2FF}
        BLE (CLE + QLE) & \textbf{85.9} & \textbf{62.2} & \textbf{80.0} & \textbf{59.7} & \textbf{53.8} & \textbf{34.9} & \textbf{62.8} & \textbf{50.6} \\
        \bottomrule
    \end{tabular}
    \end{adjustbox}
    \caption{\textbf{Ablation study on bidirectional likelihood estimation.}
    We compare the performance of each likelihood estimation: candidate likelihood estimation (CLE), query likelihood estimation (QLE), and bidirectional likelihood estimation (BLE).
    We exclude CPN in this experiment.
    }
    \label{tab:ble_full}
\end{table}
\begin{table}[!t]
    \centering
    \begin{adjustbox}{width=\linewidth}
    \begin{tabular}{l|c|c c|c c|c c|c c}
        \toprule
        & \multirow{2}{*}{\textbf{CPN}} & \multicolumn{2}{c|}{\textbf{DiDeMo}} & \multicolumn{2}{c|}{\textbf{ActivityNet}} & \multicolumn{2}{c|}{\textbf{LSMDC}} & \multicolumn{2}{c}{\textbf{MSRVTT}} \\
        & & T2V & V2T & T2V & V2T & T2V & V2T & T2V & V2T \\
        \midrule
        \midrule
        CLE & \ding{56} & \textbf{45.1} & 23.7 & 39.8 & 18.2 & 27.7 & 10.7 & \textbf{38.5} & 14.2 \\
        \rowcolor[HTML]{BFF2FF}
        CLE & \ding{52} & \textbf{45.1} & \textbf{73.3} & \textbf{41.3} & \textbf{51.3} & \textbf{28.9} & \textbf{34.5} & \textbf{38.5} & \textbf{50.0} \\
        \midrule
        BLE & \ding{56} & \textbf{85.9} & 62.2 & \textbf{80.0} & 59.7 & \textbf{53.8} & 34.9 & \textbf{62.8} & 50.6 \\
        \rowcolor[HTML]{BFF2FF}
        BLE & \ding{52} & \textbf{85.9} & \textbf{76.7} & \textbf{80.0} & \textbf{67.4} & \textbf{53.8} & \textbf{41.3} & \textbf{62.8} & \textbf{55.8} \\
        \bottomrule
    \end{tabular}
    \end{adjustbox}
    \caption{\textbf{Ablation study on CPN.}
    }
    \label{tab:cpn_full}
\end{table}
\begin{figure*}[!t]
    \centering
    \begin{subfigure}[h]{0.49\linewidth}
        \includegraphics[width=1.0\linewidth]{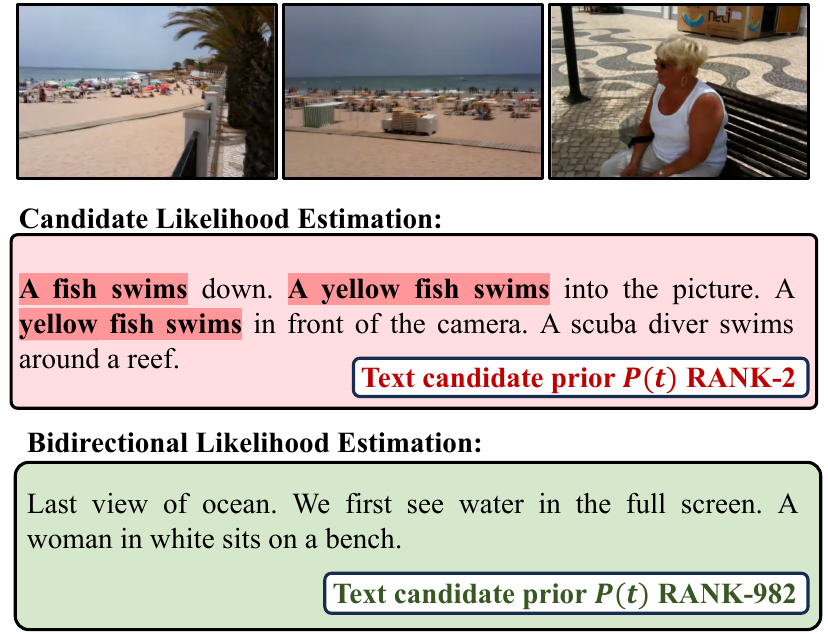}
        \caption{\textbf{Video-to-Text Retrieval.}}
        \label{fig:ble1}
    \end{subfigure}
    \begin{subfigure}[h]{0.49\linewidth}
        \includegraphics[width=1.0\linewidth]{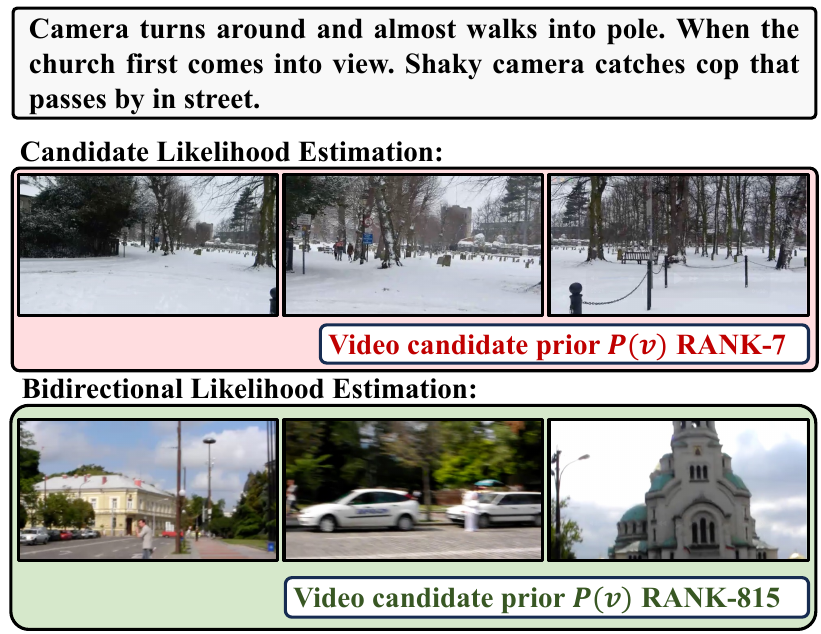}
        \caption{\textbf{Text-to-Video Retrieval.}}
        \label{fig:ble2}
    \end{subfigure}
    \caption{Qualitative results of the bidirectional likelihood estimation in (a) video-to-text and (b) text-to-video retrieval.}
    \label{fig:ble_supp}
\end{figure*}
\section{Further Qualitative Results}
\label{sup:qualitative}

\subsection{Results on Bidirectional Likelihood Estimation}
\label{sup:ble_qual}

In Fig.~\ref{fig:ble_supp}, we provide additional qualitative results on bidirectional likelihood estimation for both video-to-text and text-to-video retrieval.
We observe that candidate likelihood estimation tends to favor text and video candidates with high prior probability (ranked 2nd and 7th out of 1,003 candidates) on video-to-text (Fig.~\ref{fig:ble1}) and text-to-video (Fig.~\ref{fig:ble2}) retrieval, respectively. 
Interestingly, the high-prior text candidate contains repetitive phrases due to the autoregressive property of the LLM~\cite{wang2024mitigating}.
Likewise, the high-prior video candidate consists of static scenes, while the ground-truth video exhibits richer temporal dynamics.
However, our bidirectional likelihood estimation successfully retrieves the correct text and video in both tasks. 
These results demonstrate that candidate prior bias can lead to inaccurate retrieval, while our method effectively mitigates this bias, resulting in improved retrieval performance.

\subsection{Results on Candidate Prior Normalization}
\label{sup:cpn_qual}

\begin{figure}[!t]
    \centering
    \includegraphics[width=1.0\linewidth]{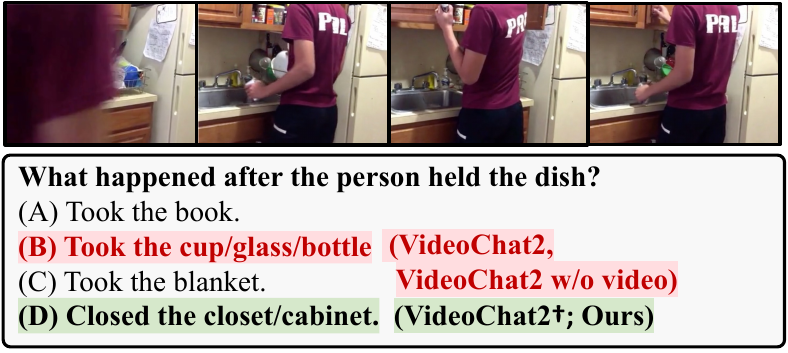}
    \caption{\textbf{A qualitative example of CPN decoding on MVBench.}
    Green signifies the accurate prediction, while red denotes the incorrect prediction.
    $\dagger$ indicates the model with CPN decoding.
    }
    \label{fig:cpn_decoding_sup}
\end{figure}

We provide further qualitative results of CPN decoding in Fig.~\ref{fig:cpn_decoding_sup} and identify a bias towards \textit{frequent co-occurrence}.
The VideoChat2 w/o video model prioritizes the likely action sequence ``(B) Took the cup/glass/bottle'' in response to the question ``What happened after the person held the dish?'', based on the frequent co-occurrence derived from the LLM's pretrained knowledge.
Consequently, the standard VideoChat2's high dependence on incorrect text priors leads to inaccurate outputs, whereas our CPN decoding effectively reduces this bias by leading the model to focus more on visual information.

\subsection{Results on Instruction-based Retrieval}
\label{sup:instruction}

\begin{figure*}[!ht] 
    \centering
    \begin{subfigure}[h]{0.95\linewidth}
        \includegraphics[width=1.0\linewidth]{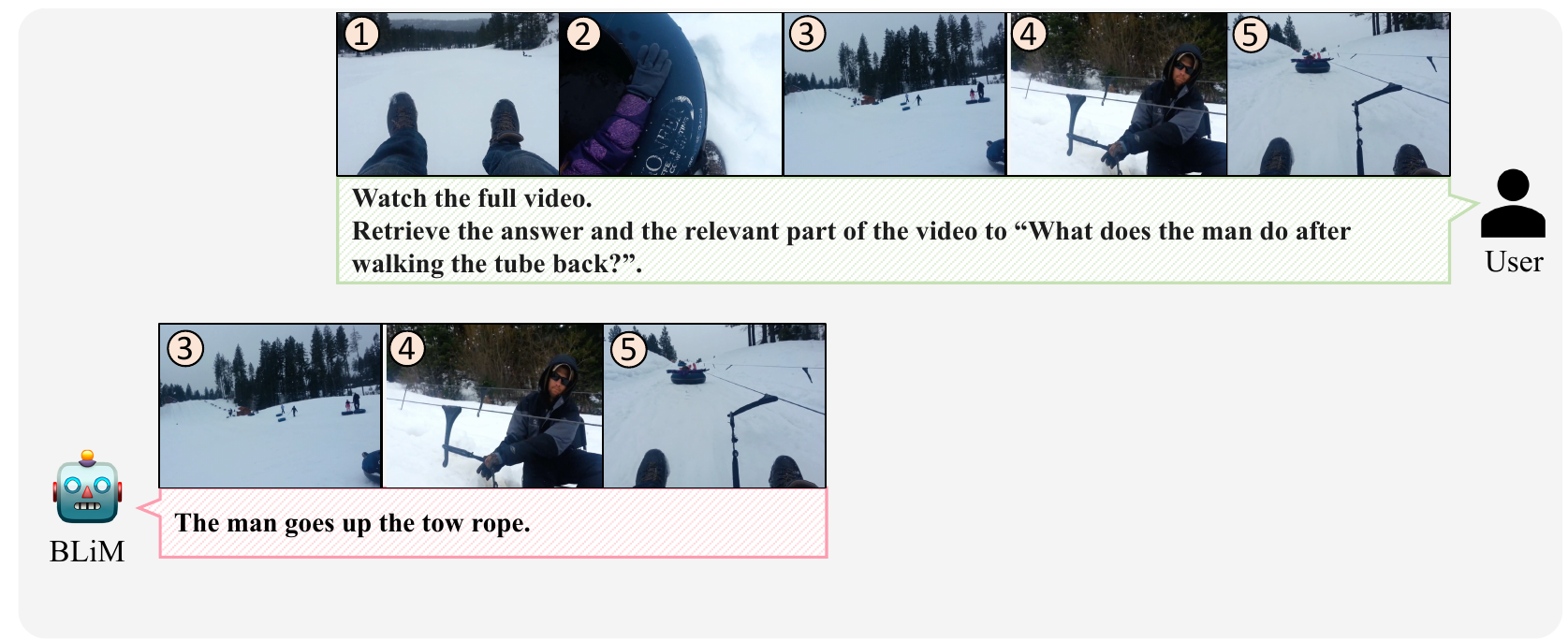}
        \caption{}
        \label{fig:instruction1}
    \end{subfigure}
    \begin{subfigure}[h]{0.95\linewidth}
        \includegraphics[width=1.0\linewidth]{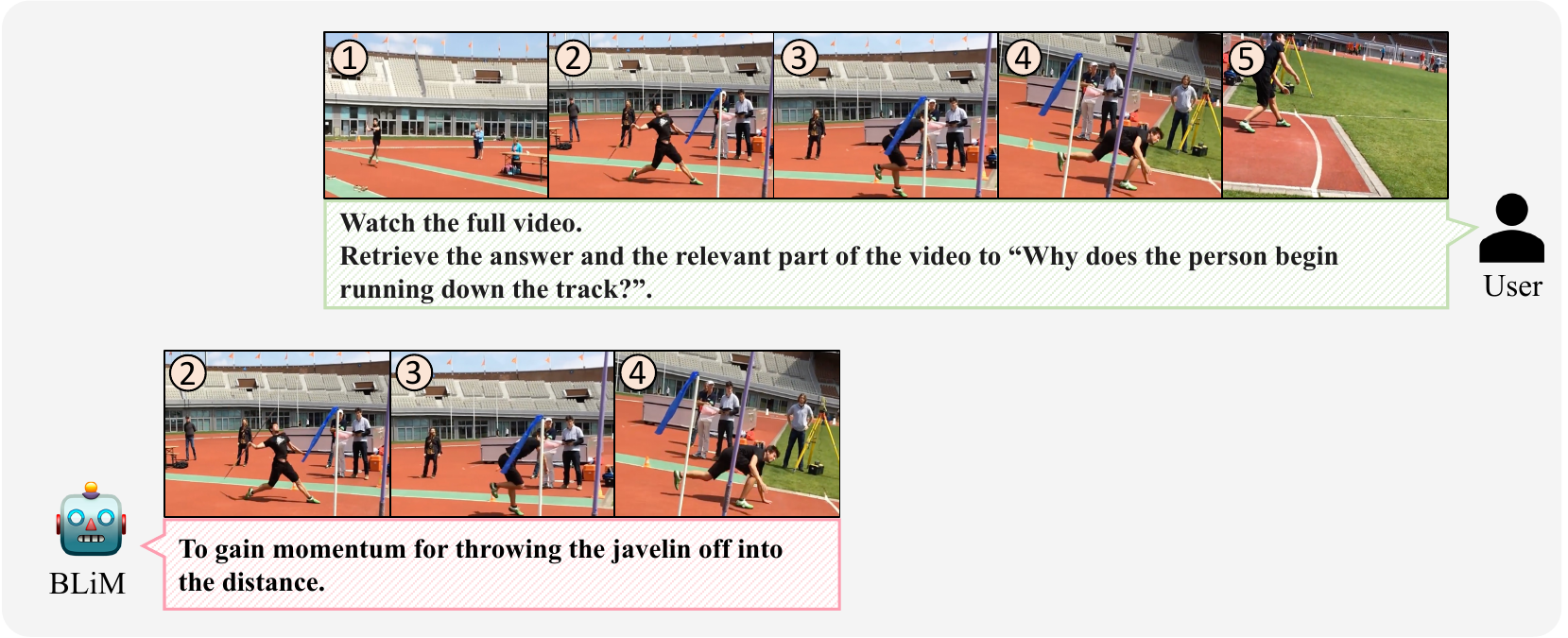}
        \caption{}
        \label{fig:instruction2}
    \end{subfigure}
    \begin{subfigure}[h]{0.95\linewidth}
        \includegraphics[width=1.0\linewidth]{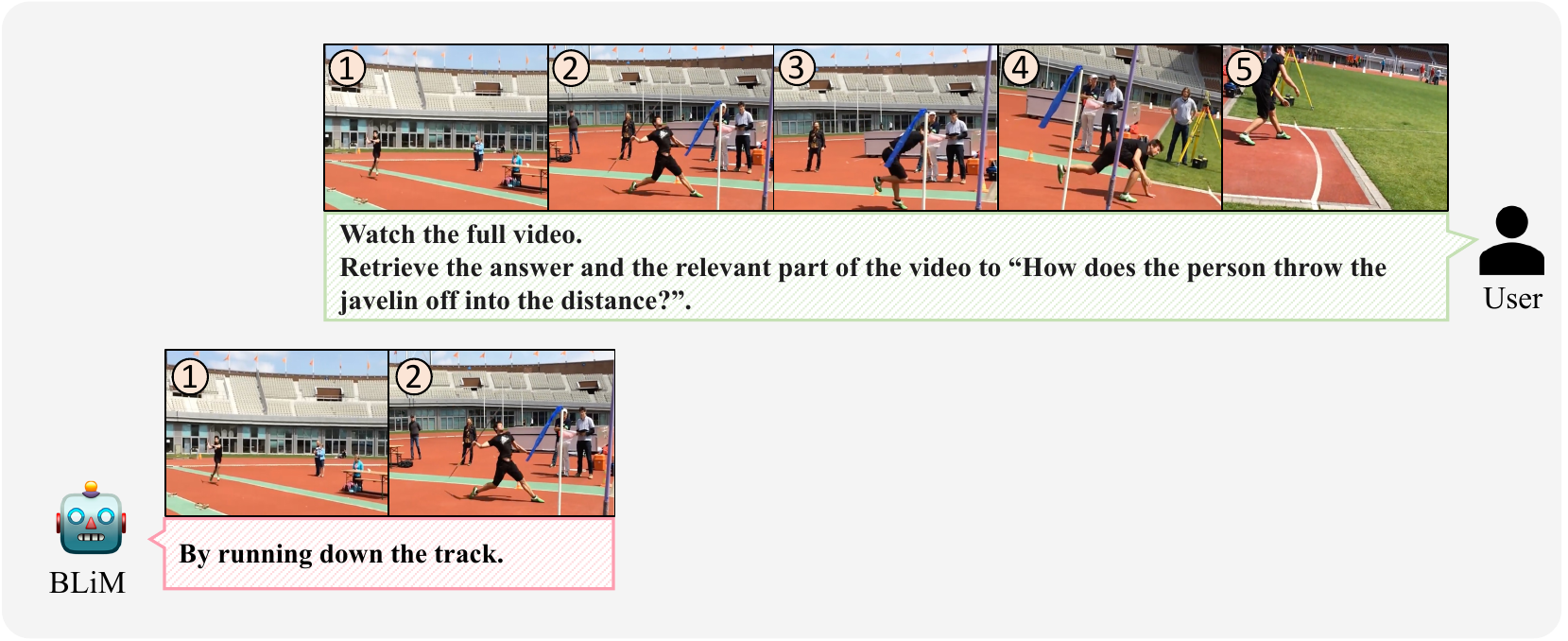}
        \caption{}
        \label{fig:instruction3}
    \end{subfigure}
    \caption{\textbf{Qualitative results of human instruction-based retrieval on ReXTime.}
    }
    \label{fig:instruction}
\end{figure*}

In this section, we explore the MLLMs' versatility in the human instruction-based retrieval task.
We note that the benchmark for human instruction-based retrieval is not yet studied, so we customize ReXTime~\citep{chen2024rextime}, originally released for the moment-retrieval task, adequately to our setting and we provide qualitative results on several examples.
In Fig.~\ref{fig:instruction}, we mainly ask the model to retrieve a certain part of the video and the answer given the video and question, \ie, multi-modal queries and multi-modal contents.
Specifically, in Fig.~\ref{fig:instruction1}, the user asks to retrieve the answer and the relevant part of the video to ``What does the man do after walking the tube back?''.
Our BLiM successfully retrieves the relevant part of the video including the 3rd, 4th, and 5th frames along with the text ``The man goes up the tow rope.'', as the action ``walking the tube back'' occurs in the 3rd frame.
This retrieved video includes the action where the man goes up the tow rope.
Furthermore, we ask two different questions with the same video in Fig.~\ref{fig:instruction2} and \ref{fig:instruction3}.
Our model retrieves the relevant part of the video and the answer well by following the instructions.
In Fig.~\ref{fig:instruction2}, the scene of gaining momentum for throwing the javelin and the text ``To gain momentum for throwing the javelin off into the distance.'' are retrieved given the question ``Why does the person begin running down the track?'' and the full video.
Interestingly, as the question is changed to ``How does the person throw the javelin off into the distance?'', the retrieved scene and text are changed to the content depicting ``running down the track''.
Overall, integrating the retrieval task into MLLMs enables them to handle complex human instruction-based retrieval in the real-world chatting system.

\newpage

\end{document}